%% file: paper.tex
\documentclass[english,runningheads,a4paper]{llncs}
\usepackage[paperheight=236mm,paperwidth=157mm,height=193mm,width=122mm]{geometry}

\input{preamble}

\usepackage{fancyhdr}

\begin{document}

\title{Optical Flow Estimation \\ in the Deep Learning Age}
\author{Junhwa Hur \and Stefan Roth}
\institute{Department of Computer Science, TU Darmstadt, Germany \\ \email{ \{junhwa.hur, stefan.roth\}@visinf.tu-darmstadt.de}}

\maketitle
\thispagestyle{fancy}

\begin{abstract}
  \input{chJH_abstract}
\end{abstract}

\input{chJH_sec1}

\input{chJH_sec2}
\input{chJH_sec3}

\input{chJH_conclusion}

\renewcommand{\bibsection}{\section*{References}} 

{
\bibliographystyle{splncsnat}
  \small 

\input{paper.bbl}
}
\ \\
\end{document}

%% file: chJH_abstract.tex
Akin to many subareas of computer vision, the recent advances in deep learning have also significantly influenced the literature on optical flow.
Previously, the literature had been dominated by classical energy-based models, which formulate optical flow estimation as an energy minimization problem.
However, as the practical benefits of Convolutional Neural Networks (CNNs) over conventional methods have become apparent in numerous areas of computer vision and beyond, they have also seen increased adoption in the context of motion estimation to the point where the current state of the art in terms of accuracy is set by CNN approaches.
We first review this transition as well as the developments from early work to the current state of CNNs for optical flow estimation. 
Alongside, we discuss some of their technical details and compare them to recapitulate which technical contribution led to the most significant accuracy improvements.
Then we provide an overview of the various optical flow approaches introduced in the deep learning age, including those based on alternative learning paradigms (\eg, unsupervised and semi-supervised methods) as well as the extension to the multi-frame case, which is able to yield further accuracy improvements.

%% file: chJH_sec1.tex
\section{Emergence and Advances of Deep Learning-based Optical Flow Estimation}
\label{chJH_sec:1}

The recent advances in deep learning have significantly influenced the literature on optical flow estimation and fueled a transition from classical energy-based formulations, which were mostly hand defined, to end-to-end trained models.
We first review how this transition proceeded by recapitulating early work that started to utilize deep learning, typically as one of several components.
Then, we summarize several canonical end-to-end approaches that have successfully adopted CNNs for optical flow estimation and have highly influenced the mainstream of research, including other subareas of vision in which optical flow serves as an input.

\subsection{From classical energy-based approaches to CNNs}
\label{chJH_subsec:1_1}

\begin{figure}[t]
    \centering
    \begin{subfigure}[t]{0.45\textwidth}
        \includegraphics[width=\textwidth]{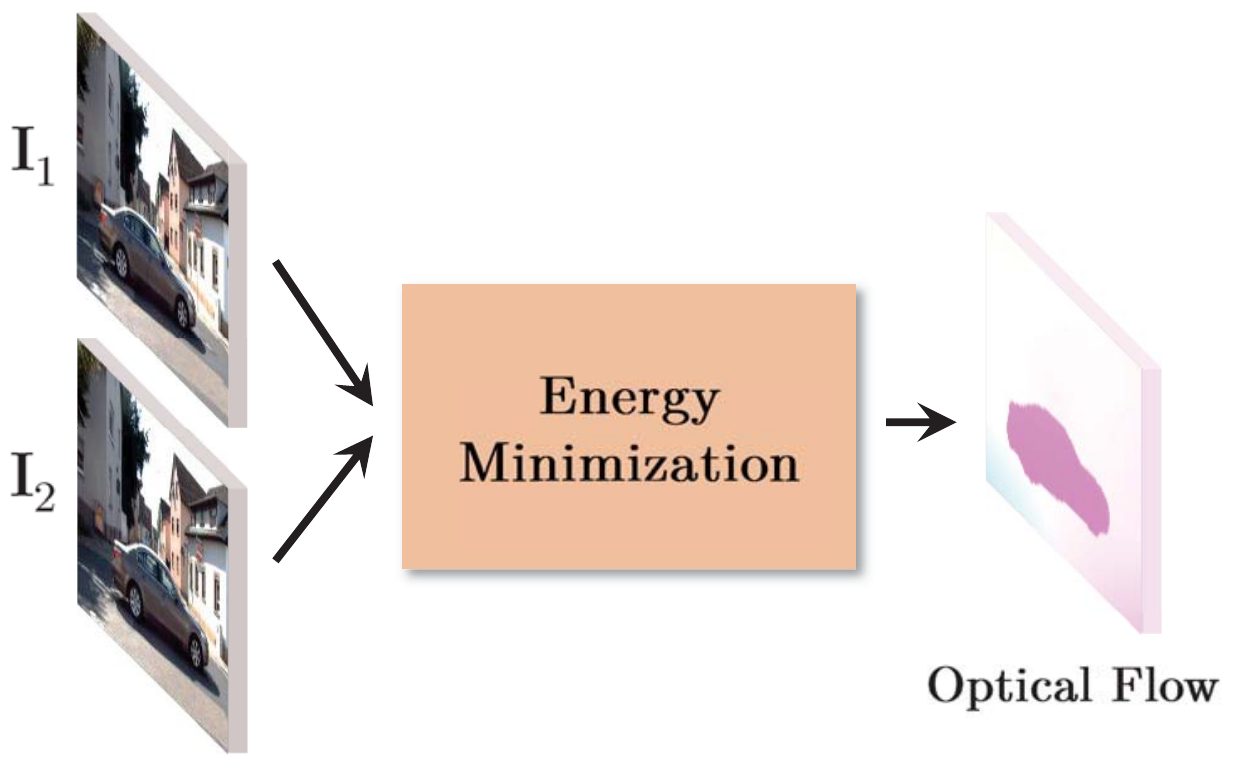}
        \caption{Classical energy-based approach}
        \label{chJH:fig:energy}
    \end{subfigure}
    \\[1mm]
    \begin{subfigure}[t]{0.51\textwidth}
        \includegraphics[width=\textwidth]{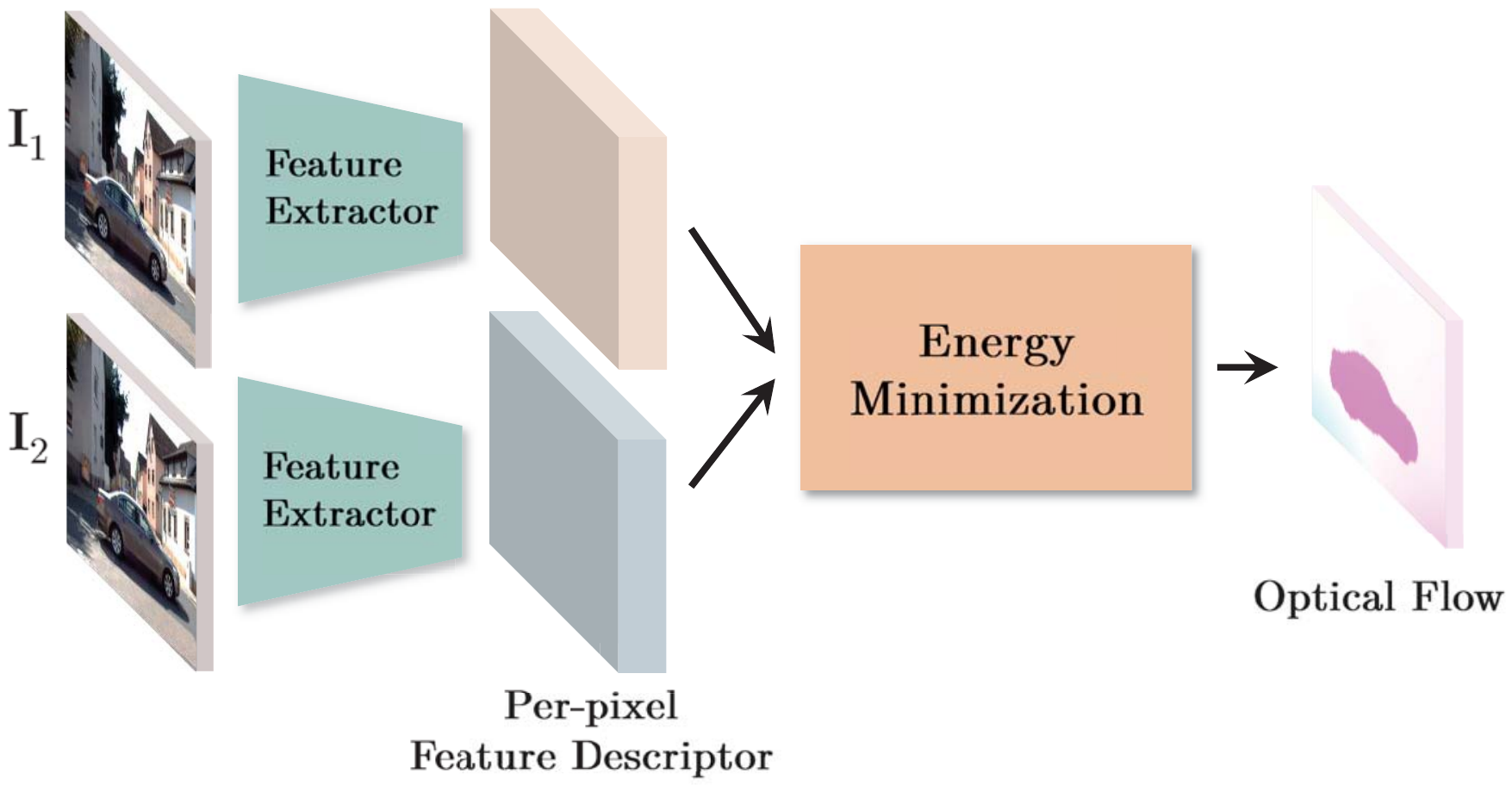}
        \caption{Using CNNs as a feature extractor}
        \label{chJH:fig:feature}
    \end{subfigure}
    \quad
    \begin{subfigure}[t]{0.45\textwidth}
        \includegraphics[width=\textwidth]{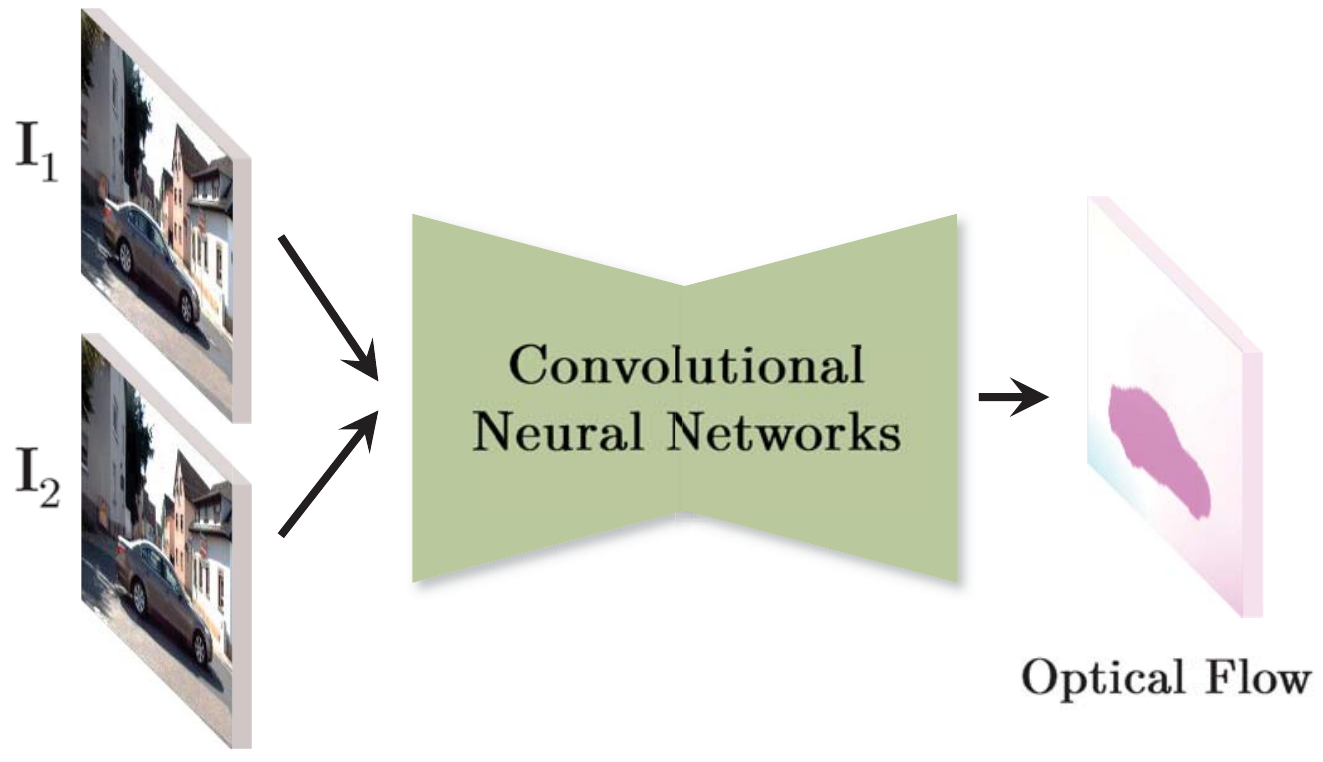}
        \caption{CNN regression architecture}
        \label{chJH:fig:generic}
    \end{subfigure}
    \caption{Transition from \emph{(a)} classical energy-based approaches to \emph{(b)} CNN-based approaches that use CNNs as a feature extractor or to \emph{(c)} end-to-end trainable CNN regression architectures.}\label{chJH:fig:energy_and_cnns}
\end{figure}

For more than three decades, research on optical flow estimation has been heavily influenced by the variational approach of Horn and Schunck \cite{chJH:Horn:1981:DOF}. 
Their basic energy minimization formulation consists of a data term, which encourages brightness constancy between temporally corresponding pixels, and a spatial smoothness term, which regularizes neighboring pixels to have similar motion in order to overcome the aperture problem.
The spatially continuous optical flow field $\textbf{u}=(u_x, u_y)$ is obtained by minimizing
\begin{equation}
E(\textbf{u}) = \int \Big( (I_x u_x + I_y u_y + I_t)^2 + \alpha^2\big(\lVert \nabla u_x \rVert^2 + \lVert \nabla u_y \rVert^2 \big) \Big) \;dx \,dy,
\label{chJH:eq:Horn_Schunck} 
\end{equation}  
where $I_x, I_y, I_t$ are the partial derivatives of the image intensity $I$ with respect to $x$, $y$, and $t$ (Fig.~\ref{chJH:fig:energy}).
To minimize Eq.~\eqref{chJH:eq:Horn_Schunck} in practice, spatial discretization is necessary.
In such a spatially discrete form, the Horn and Schunck model \cite{chJH:Horn:1981:DOF} can also be re-written in the framework of standard pairwise Markov random fields (MRFs) \cite{chJH:Boykov:1998:MRF,chJH:Li:1994:MRF} through a combination of a unary data term $D(\cdot)$ and a pairwise smoothness term $S(\cdot, \cdot)$,
\begin{equation}
E(\textbf{u}) = \sum_{\textbf{p} \in \mathcal{I}} D(\textbf{u}_\textbf{p}) + \sum_{\textbf{p}, \textbf{q} \in \mathcal{N}} S(\textbf{u}_\textbf{p}, \textbf{u}_\textbf{q}),
\label{chJH:eq:Horn_Schunck_mrf} 
\end{equation}  
where $\mathcal{I}$ is the set of image pixels and the set $\mathcal{N}$ denotes spatially neighboring pixels.
Starting from this basic formulation, much research has focused on designing better energy models that more accurately describe the flow estimation problem (see \cite{Fortun:2015:OFM,Tu:2019:ASV} for reviews of such methods).

Concurrently with pursing better energy models, the establishment of public benchmark datasets for optical flow, such as the Middlebury \cite{chJH:Baker:2011:DBE}, MPI Sintel \cite{chJH:Butler:2012:NOS}, and KITTI Optical Flow benchmarks \cite{chJH:Geiger:2012:AWR,chJH:Menze:2015:OSF}, has kept revealing the challenges and limitations of existing methods.
These include large displacements, severe illumination changes, and occlusions.
Besides allowing for the fair comparison between existing methods on the same provided data, these public benchmarks have moreover stimulated research on more faithful energy models that address some of the specific challenges mentioned above.

Meanwhile, the relatively recent success of applying Convolutional Neural Networks (CNNs) with backpropagation on a large-scale image classification task \cite{chJH:Krizhevsky:2013:ICD} paved the way for applying CNNs to various other computer vision problems, including optical flow as well.
Early work that applied CNNs to optical flow used them as an advanced feature extractor \cite{chJH:Bai:2016:ESI,chJH:Bailer:2017:CPM,chJH:Gadot:2016:PBB,chJH:Gueney:2016:DDF}, as sketched in Fig.~\ref{chJH:fig:feature}.
The main idea behind this is to substitute the data term (\eg,~in Eqs.~\eqref{chJH:eq:Horn_Schunck} and \eqref{chJH:eq:Horn_Schunck_mrf}) in classical energy-based formulations with a CNN-based feature matching term.
Instead of using image intensities, image gradients, or other hand-crafted features as before, CNNs enable learning feature extractors such that each pixel can be represented with a high-dimensional feature vector that combines a suitable amount of distinctiveness and invariance, for example to appearance changes.
The putative similarity between regions is given by the feature distance.
The remaining pipeline, including using the smoothness term as well as the optimization strategies, remain the same.
As we will review in more detail below, several methods \cite{chJH:Bai:2016:ESI,chJH:Bailer:2017:CPM,chJH:Gadot:2016:PBB,chJH:Gueney:2016:DDF} demonstrated an accuracy benefit of such CNN-based feature extractors.

At the same time, another line of research investigated regression-based CNN architectures that can directly estimate optical flow from a pair of input images and can be trained end-to-end, as sketched in Fig.~\ref{chJH:fig:generic}.
Unlike methods that combine CNN feature extractors with classical regularizers and energy minimization, such regression frameworks employ CNNs for the entire pipeline by virtue of their ability to act as a function approximator, which effectively learns the relationship between the input images and the desired flow output given the labeled training dataset.
FlowNet \cite{chJH:Dosovitskiy:2015:FLO} is the first work that demonstrated an end-to-end CNN regression approach for estimating optical flow based on an encoder-decoder architecture. 
Owing to the difficulty of obtaining dense ground truth optical flow in real-world images, Dosovitskiy~\etal~\cite{chJH:Dosovitskiy:2015:FLO} generated a synthetic dataset from CAD models of chairs, which move in front of a static background. 
Pairs of images with ground truth optical flow serve to train the network.
FlowNet \cite{chJH:Dosovitskiy:2015:FLO} demonstrated that a CNN-based regression architecture is able to predict optical flow directly, yet the accuracy remained behind that of state-of-the-art energy-based methods at the time \cite{chJH:Revaud:2015:EEP,chJH:Weinzaepfel:2013:DLD}.
Unlike in other areas of computer vision, this left it initially unclear whether end-to-end CNN architectures can compete with classical energy-based methods in terms of accuracy.

However, later research cleared up this question by developing better end-to-end architectures that eventually outperformed classical energy-based methods, reaching new accuracy levels on public benchmarks \cite{chJH:Butler:2012:NOS,chJH:Geiger:2012:AWR,chJH:Menze:2015:OSF}.
These advances mainly stem from discovering new architecture designs, for example, by stacking multiple networks to refine previous estimates \cite{chJH:Ilg:2017:FN2} or constructing a CNN pyramid to estimate flow in a coarse-to-fine fashion \cite{chJH:Hui:2018:LFN,chJH:Ranjan:2017:OFE,chJH:Sun:2018:PCO}, as had been done in classical methods before.
Unlike energy-based models, CNN regressors run in real time on GPUs combined with much better accuracy.
In other words, end-to-end CNN regressors have established themselves by now as dominant paradigm in the current literature on optical flow estimation.
Yet, they have not remained without limitations, hence much research continues to be carried out.
For example, recent work aims to overcome the reliance on large amounts of labeled data as well as accuracy drops on unseen domains and datasets, for example by pursuing unsupervised or semi-supervised learning paradigms.

In the following, we will give a detailed overview of the two major CNN paradigms in optical flow estimation and survey other recent trends.

\subsection{CNNs as feature extractor}
\label{chJH_subsec:1_2}

Not restricted to the problem domain of optical flow estimation but rather correspondence estimation more generally, several early works \cite{chJH:Han:2015:MUF,chJH:Simo-Serra:2015:DLD,chJH:Zbontar:2015:CSM,chJH:Zagoruyko:2015:LCI} employed CNNs for matching descriptors or patches.
In most cases, the underlying network uses a so-called Siamese architecture that extracts a learned feature descriptor separately for each of two input image patches, followed by a shallow joint network that computes a matching score between the two feature representations.
The name Siamese alludes to the fact that the two feature extractor sub-networks are identical including their weights.
Inspired by these successes, significant amounts of earlier work that adopted deep learning for optical flow estimation focused on utilizing CNNs as a feature extractor on top of conventional energy-based formulations such as MRFs.
Their main idea is to utilize CNNs as a powerful tool for extracting discriminative features and then use well-proven conventional energy-based frameworks for regularization.

Gadot and Wolf \cite{chJH:Gadot:2016:PBB} proposed a method called \textbf{PatchBatch}, which was among the first flow approaches to adopt CNNs for feature extraction.
PatchBatch \cite{chJH:Gadot:2016:PBB} is based on a Siamese CNN feature extractor that is fed $51 \times 51$ input patches and outputs a $512$-dimensional feature vector using a shallow 5-layer CNN.
Then, PatchBatch \cite{chJH:Gadot:2016:PBB} adopts Generalized PatchMatch \cite{chJH:Barnes:2010:GPC} as an Approximate Nearest Neighbor (ANN) algorithm for correspondence search, \ie,~matching the extracted features between two images.
The method constructs its training set by collecting positive corresponding patch examples given ground-truth flow and negative non-matching examples by randomly shifting the image patch in the vicinity of where the ground-truth flow directs.
The intuition of collecting negative examples in such a way is to train CNNs to be able to separate non-trivial cases and extract more discriminative features.
The shallow CNNs are trained using a variant of the DrLIM \cite{chJH:Hadsell:2006:DRL} loss, which minimizes the squared $L_2$ distance between positive patch pairs and maximizes the squared $L_2$ distance between negative pairs above a certain margin.

In a similar line of work, Bailer~\etal~\cite{chJH:Bailer:2017:CPM} proposed to use the thresholded hinge embedding loss for training the feature extractor network.
The hinge embedding loss based on the $L_2$ loss function has been commonly used to minimize the feature distance between two matching patches and to maximize the feature distance above $m$ between non-matching patches:
\begin{eqnarray}
	\label{chJH:eq:hingeloss_a}
	l_{\text{hinge}}(\mathbf{P}_1, \mathbf{P}_2) &=& \left\{ 
	\begin{array}{lll}
		L_2(\mathbf{P}_1, \mathbf{P}_2), &&(\mathbf{P}_1, \mathbf{P}_2) \in M^+ \\  
		\text{max}\big(0, m-L_2(\mathbf{P}_1, \mathbf{P}_2)\big), &&(\mathbf{P}_1, \mathbf{P}_2) \in M^-  \\
	\end{array}		
	\right. \\
	\label{chJH:eq:hingeloss_b} 
	L_2(\mathbf{P}_1, \mathbf{P}_2) &=& \big\lVert F(\mathbf{P}_1) - F(\mathbf{P}_2) \big\rVert_2,
\end{eqnarray}
where $F(\mathbf{P}_1)$ and $F(\mathbf{P}_2)$ are the extracted descriptors from CNNs applied to $\mathbf{P}_1$ in the first image and $\mathbf{P}_2$ in the second image, respectively, $L_2(\mathbf{P}_1, \mathbf{P}_2)$ calculates the $L_2$ loss between the two descriptors, and $M^+$ and $M^-$ are collected sets of positive and negative samples, respectively.

However, minimizing the $L_2$ loss of some challenging positive examples (\eg, with appearance difference or illumination changes) can move the decision boundary into an undesired direction and lead to misclassification near the decision boundary.
Thus, Bailer~\etal~\cite{chJH:Bailer:2017:CPM} proposed to put a threshold $t$ on the hinge embedding loss in order to prevent the network from minimizing the $L_2$ distance too aggressively:
\begin{equation}
	l_{t\text{-hinge}}(\mathbf{P}_1, \mathbf{P}_2) = \left\{ 
	\begin{array}{lll}
		\text{max}\big(0, L_2(\mathbf{P}_1, \mathbf{P}_2)-t\big), &&(\mathbf{P}_1, \mathbf{P}_2) \in M^+ \\  
		\text{max}\big(0, m-(L_2(\mathbf{P}_1, \mathbf{P}_2)-t)\big), &&(\mathbf{P}_1, \mathbf{P}_2) \in M^- \label{chJH:eq:th_hingeloss_a}. \\
	\end{array}		
	\right.
\end{equation}  
Compared to standard losses, such as the hinge embedding loss in Eq.~\eqref{chJH:eq:hingeloss_a} or the DrLIM loss \cite{chJH:Hadsell:2006:DRL}, this has led to more accurate flow estimates.

Meanwhile, G{\"u}ney and Geiger~\cite{chJH:Gueney:2016:DDF} demonstrated successfully combining a CNN feature matching module with a discrete MAP estimation approach based on a pairwise Markov random field (MRF) model. 
The proposed CNN module outputs per-pixel descriptors, from which a cost volume is constructed by calculating feature distances between sample matches.
This is input to a discrete MAP estimation approach \cite{chJH:Menze:2015:DOO} to infer the optical flow.
To keep training efficient, G{\"u}ney and Geiger~\cite{chJH:Gueney:2016:DDF} followed a piece-wise setting that first trains the CNN module alone and only then trains the joint CNN-MRF module together.
Bai \etal~\cite{chJH:Bai:2016:ESI} followed a similar setup overall, but utilized semi-global block matching (SGM) \cite{Hirschmuller:2008:SPS} to regress the output optical flow from the cost volume, which is constructed by calculating a distance between features from CNNs.

Taken together, these approaches have successfully demonstrated that the benefits of the representational power of CNNs can be combined with well-proven classical energy-based models. 
Specifically, they demonstrated more accurate estimates on inliers and more precise estimates on object boundaries than previous baselines with hand-constructed features.

\subsection{End-to-end regression architectures for optical flow estimation}
\label{chJH_subsec:1_3}

Concurrently with the development of feature extraction-based networks, active research also started on developing end-to-end CNN architectures for optical flow estimation based on regression.
Unlike methods that use CNNs only for feature extraction as addressed above, such regression methods exploit CNNs for the entire pipeline and directly output optical flow from a pair of input images.
By substituting classical regularizers and avoiding energy minimization, these CNN-based methods combine the advantages of end-to-end trainability and runtime efficiency.

Dosovitskiy~\etal~ proposed the first end-to-end CNN architecture for estimating optical flow, called FlowNet \cite{chJH:Dosovitskiy:2015:FLO}, which has two main architectural lines, \textbf{FlowNetS} and \textbf{FlowNetC}.
The two models are fundamentally based on an hourglass-shaped neural network architecture that consists of an encoder and a decoder, and differs only in the encoder part.
In FlowNetS, a pair of input images is simply concatenated and then input to the hourglass-shaped network that directly outputs optical flow.
On the other hand, FlowNetC has a shared encoder for both images, which extracts a feature map for each input image, and a cost volume is constructed by measuring patch-level similarity between the two feature maps with a correlation operation.
The result is fed into the subsequent network layers.

To train the networks in a supervised way, a training dataset with a large number of image pairs and their ground truth flow are required, but at the time only datasets with few hundreds of images or even fewer were available \cite{chJH:Baker:2011:DBE,chJH:Geiger:2012:AWR,chJH:Menze:2015:OSF};  the challenge of obtaining dense optical flow ground truth in real-world images remains until today.
In order to overcome the shortage of suitable training data, Dosovitskiy~\etal~\cite{chJH:Dosovitskiy:2015:FLO} established a synthetic dataset, called FlyingChairs, by layering natural images with rendered CAD models of chairs; their parameterized affine motion is designed to follow the motion statistics of existing real-world datasets.
However, due to the intrinsic differences between synthetic and real-world images, unfortunately FlowNet trained on the synthetic dataset alone did not generalize well to real images.
In fact, even after fine-tuning on real-world images, the accuracy initially remained behind that of classical energy-based models at the time.
This left the question whether such a generic CNN regression architecture can actually outperform classical energy-based methods, or why it did not (yet).
Importantly, however, FlowNet \cite{chJH:Dosovitskiy:2015:FLO} demonstrated the possibility of employing an end-to-end regression architecture for optical flow estimation.
Moreover, FlowNet established several standard practices for training optical flow networks such as learning rate schedules, basic network architectures, data augmentation schemes, and the necessity of pre-training on synthetic datasets, which have substantially impacted follow-up research.

\begin{figure}[t]
    \centering
    \begin{subfigure}[t]{0.49\textwidth}
        \includegraphics[width=\textwidth]{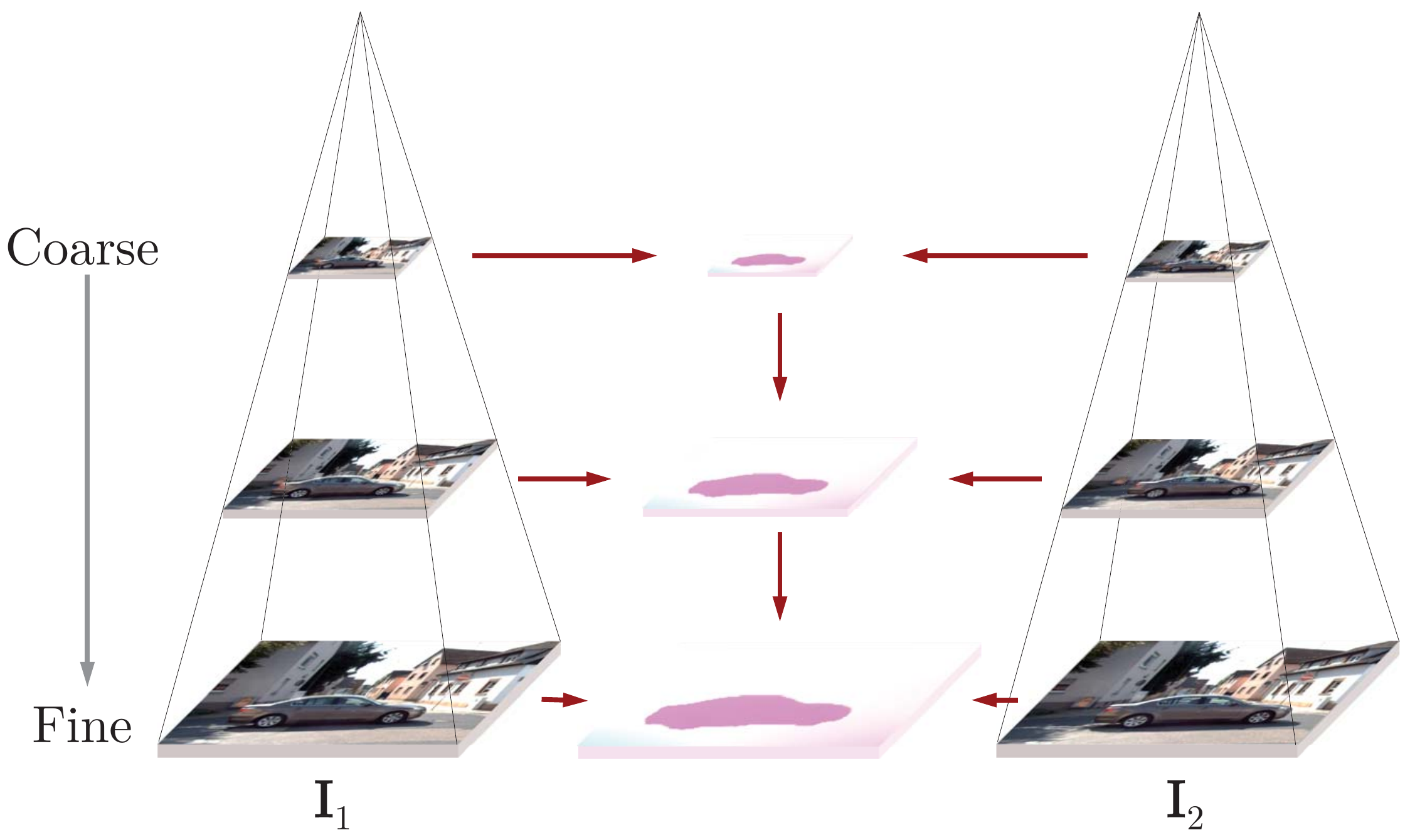}
        \caption{Coarse-to-fine estimation}
        \label{chJH:fig:coarse2fine}
    \end{subfigure}
    \quad \quad
    \begin{subfigure}[t]{0.44\textwidth}
    	\raisebox{6mm}{
        \includegraphics[width=\textwidth]{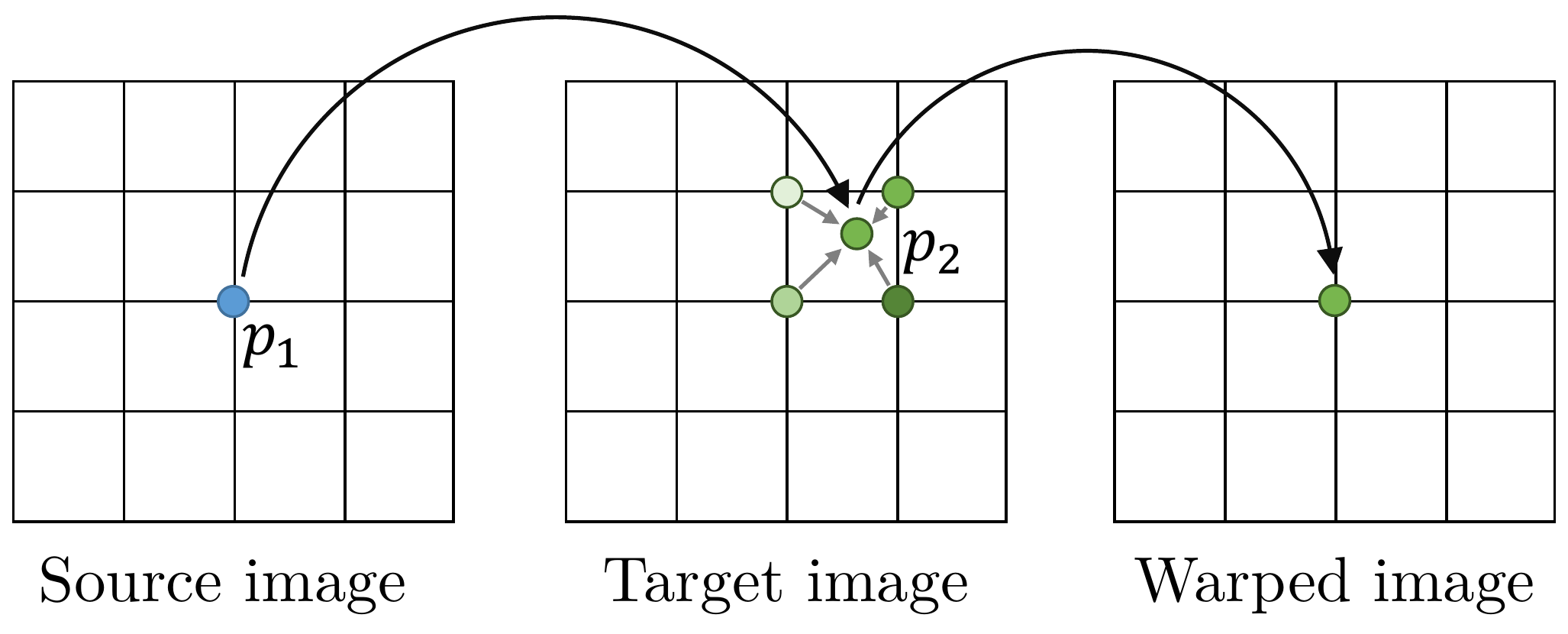}
        }
        \caption{Backward warping}
        \label{chJH:fig:warping}        
    \end{subfigure}
    \quad
    \caption{\emph{(a)} The classical coarse-to-fine concept proceeds by estimating optical flow using a multi-scale image pyramid, starting from the coarsest level to the finest level. 
    By gradually estimating and refining optical flow through the pyramid levels, this approach can handle large displacements better and improve accuracy. 
    \emph{(b)} Backward warping is commonly used in optical flow estimation.
    For each pixel $p_1$ in the source image, the warped image obtains the intensity from (sub)pixel location $p_2$, which is obtained from the estimated flow. 
    Bilinear interpolation is often used to obtain the pixel intensity at the non-integer coordinate.}
    \label{chJH:fig:coarse2fine_warping}
\end{figure}


Ranjan~\etal~proposed \textbf{SPyNet} \cite{chJH:Ranjan:2017:OFE}, which incorporates the classical ``coarse-to-fine'' concept (please refer to Fig.~\ref{chJH:fig:coarse2fine} for an illustration) into a CNN model and updates the residual flow over multiple pyramid levels.
SPyNet consists of 5 pyramid levels, and each pyramid level consists of a shallow CNN that estimates flow between a source image and a target image, which is warped by the current flow estimate (see Fig.~\ref{chJH:fig:warping}).
This estimate is updated so that the network can residually refine optical flow through a spatial pyramid and possibly handle large displacements.
Compared to FlowNet, SPyNet significantly reduces the number of model parameters by 96\% by using a pyramid-shaped architecture, while achieving comparable and sometimes even better results than FlowNet.
Although SPyNet \cite{chJH:Ranjan:2017:OFE} is still outperformed by classical energy-based methods, it demonstrates a promising way of designing flow architectures by integrating classical principles into deep learning.

Meanwhile, Ilg~\etal~\cite{chJH:Ilg:2017:FN2} proposed \textbf{FlowNet2}, which significantly improves the flow accuracy over their previous FlowNet architecture and started to outperform classical energy-based approaches.
The main limitations of FlowNet are blurry outputs from the CNN decoder and lower accuracy compared to classical approaches.
To overcome these limitations, Ilg~\etal~proposed the key idea that by stacking multiple FlowNet-style networks, one can sequentially refine the output from the previous network modules.
Despite of the conceptual simplicity, stacking multiple networks is very powerful and significantly improves the flow accuracy by more than 50\% over FlowNet.
Additionally, Ilg~\etal~revealed several important practices for training their networks, including the necessity of pre-training and fine-tuning on synthetic datasets, the effectiveness of using a correlation layer, and the guidance of proper learning rate schedules, followed by in-depth empirical analyses.
In practice, Ilg~\etal~\cite{chJH:Ilg:2017:FN2} suggest to pre-train their networks on a less challenging synthetic dataset first (\ie,~the FlyingChairs dataset \cite{chJH:Dosovitskiy:2015:FLO}) and then further train on a more challenging synthetic dataset with 3D motion and photometric effects (\ie,~the FlyingThings3D dataset \cite{chJH:Mayer:2016:ALD}).
Their empirical study revealed a more than 20\% accuracy difference depending on the usage of the proper pre-training dataset (see Table 1 in \cite{chJH:Ilg:2017:FN2}).
The underlying conjecture is that making the network first learn the general concept of motion estimation with a simpler dataset is more important than learning to handle various challenging examples from the start. 
Also, the proposed learning rate schedules for pre-training and fine-tuning have become a standard and guidance for follow-up research.

After the successful demonstration of FlowNet2 \cite{chJH:Ilg:2017:FN2} that  end-to-end regression architectures can outperform energy-based approaches, further investigations on finding better network architectures have continued.
Sun~\etal~proposed an advanced architecture called \textbf{PWC-Net}~\cite{chJH:Sun:2018:PCO} by exploiting well-known design principles from classical approaches.
PWC-Net relies on three main design principles: \emph{(i)} pyramid, \emph{(ii)} warping, and \emph{(iii)} cost volume.
Similar to SPyNet \cite{chJH:Ranjan:2017:OFE}, PWC-Net estimates optical flow in a coarse-to-fine way with several pyramid levels, but PWC-Net constructs a feature pyramid by using CNNs, while SPyNet constructs an image pyramid by simply downsampling images.
Next, PWC-Net constructs a cost volume with a feature map from the source image and the warped feature map from the target image based on the current flow.
Then, the subsequent CNN modules act as a decoder that outputs optical flow from the cost volume.
In terms of both accuracy and practicality, PWC-Net \cite{chJH:Sun:2018:PCO} set a new state of the art with its light-weight architecture allowing for shorter training times, faster inference, and more importantly, clearly improved accuracy.
Comparing to FlowNet2 \cite{chJH:Ilg:2017:FN2}, PWC-Net is 17 times smaller in model size and twice as fast during inference while being more accurate. 
Similar to SPyNet, the computational efficiency stems from using coarse-to-fine estimation, but PWC-Net crucially demonstrates that constructing and warping feature maps instead of using downsampled warped images yields much better accuracy.

As a concurrent work and similar to PWC-Net \cite{chJH:Sun:2018:PCO}, \textbf{LiteFlowNet} \cite{chJH:Hui:2018:LFN} also demonstrated utilizing a multi-level pyramid architecture that estimates flow in a coarse-to-fine manner, proposing another light-weight regression architecture for optical flow.
The major technical differences to PWC-Net are that LiteFlowNet residually updates optical flow estimates over the pyramid levels and proposes a flow regularization module.
The proposed flow regularization module creates per-pixel local filters using CNNs and applies the filters to each pixel so that customized filters refine flow fields by considering neighboring estimates.
The regularization module is given the optical flow, feature maps, and occlusion probability maps as inputs to take motion boundary information and occluded areas into account in creating per-pixel local filters.
The experimental results demonstrate clear benefits, especially from using the regularization module that smoothes the flow fields while effectively sharpening motion boundaries, which reduces the error by more than 13\% on the training domain.

Afterwards, Hur and Roth \cite{chJH:Hur:2019:IRR} proposed an iterative estimation scheme with weight sharing entitled \emph{iterative residual refinement} (\textbf{IRR}), which can be applied to several backbone architectures and improves the accuracy further.
Its main idea is to take the output from a previous pass through the network as input and iteratively refine it by only using a single network block with shared weights; this allows the network to residually refine the previous estimate.
The IRR scheme can be used on top of various flow architectures, for example FlowNet~\cite{chJH:Dosovitskiy:2015:FLO} and PWC-Net~\cite{chJH:Sun:2018:PCO}. 
For FlowNet~\cite{chJH:Dosovitskiy:2015:FLO}, the whole hourglass shape network is iteratively re-used to keep refining its previous estimate and, in contrast to FlowNet2~\cite{chJH:Ilg:2017:FN2}, increases the accuracy without adding any parameters.
For PWC-Net~\cite{chJH:Sun:2018:PCO}, a repetitive but separate flow decoder module at each pyramid level is replaced with only one common decoder for all levels, and then iteratively refines the estimation through the pyramid levels.
Applying the scheme on top of PWC-Net~\cite{chJH:Sun:2018:PCO} is more interesting as it makes an already lean model even more compact by removing repetitive modules that perform the same functionality.
Yet, the accuracy is improved, especially on unseen datasets (\ie~allowing better generalization).
Furthermore, Hur and Roth \cite{chJH:Hur:2019:IRR} also demonstrated an extension to joint occlusion and bi-directional flow estimation that leads to further flow accuracy improvements of up to 17.7\% while reducing the number of parameters by 26.4\% in case of PWC-Net; this model is termed \textbf{IRR-PWC} \cite{chJH:Hur:2019:IRR}.

Yin~\etal~\cite{chJH:Yin:2019:HDD} proposed a general probabilistic framework termed \textbf{HD$^3$} for dense pixel correspondence estimation, exploiting the concept of the so-called match density, which enables the joint estimation of optical flow and its uncertainty.
Mainly following the architectural design of PWC-Net (\ie, using a multi-scale pyramid, warping, and a cost volume), the method estimates the full match density in a hierarchical and computationally efficient manner.
The estimated spatially discretized match density can then be converted into optical flow vectors while providing an uncertainty assessment at the same time.
This output representation of estimating the match density is rather different from all previous works above, which directly regress optical flow with CNNs. 
On established benchmarks datasets, their experimental results demonstrate clear advantages, achieving state-of-the-art accuracy regarding both optical flow and uncertainly measures.

\begin{table}[t]
{
\setlength\tabcolsep{1.9pt} 
\centering
\caption{Overview of the main technical design principles of end-to-end optical flow architectures.}
\label{chJH:tab:genericCNN}
\scriptsize
\begin{tabularx}{\textwidth}{@{}X*{8}{c}@{}}
\toprule
Methods & \rotatebox{60}{FlowNetS \cite{chJH:Dosovitskiy:2015:FLO}} & \rotatebox{60}{FlowNetC \cite{chJH:Dosovitskiy:2015:FLO}} & \rotatebox{60}{SPyNet \cite{chJH:Ranjan:2017:OFE}} & \rotatebox{60}{FlowNet2 \cite{chJH:Ilg:2017:FN2}} & \rotatebox{60}{PWC-Net \cite{chJH:Sun:2018:PCO}} & \rotatebox{60}{LiteFlowNet \cite{chJH:Hui:2018:LFN}} & \rotatebox{60}{HD$^3$ \cite{chJH:Yin:2019:HDD}} & \rotatebox{60}{VCN \cite{chJH:Yang:2019:VCN}} \\
\midrule
Pyramid & -- & \makecell{3-level \\ feature} & \makecell{5-level \\ image} & \makecell{3-level \\ feature} & \makecell{6-level \\ feature} & \makecell{6-level \\ feature} & \makecell{5-level \\ feature} & \makecell{6-level \\ feature} \\[7pt]
Warping & -- & -- & Image & Image & Feature & Feature & Feature & Feature \\[7pt] 
Cost volume & -- & 3D & -- & 3D & 3D & 3D & 3D & 4D \\[7pt]
Network stacking & -- & -- & -- & 5 & -- & -- & -- & -- \\[4pt]
Flow inference & Direct & Direct & Residual & Direct & Direct & Residual & Residual & \makecell{Hypothesis \\ selection} \\[7pt]
Parameters (M) & 38.67 & 39.17 & 1.20 & 162.49 & 8.75 & 5.37 & 39.6 & 6.20 \\
\bottomrule
\end{tabularx}
}
\end{table}


While the cost volume has been commonly used in backbone architectures \cite{chJH:Dosovitskiy:2015:FLO,chJH:Hui:2018:LFN,chJH:Ilg:2017:FN2,chJH:Sun:2018:PCO,chJH:Yin:2019:HDD}, its representation is mainly based on a heuristic design.
Instead of representing the matching costs between all pixels $(x, y)$ with their possible 2D displacements $(u, v)$ into a 4D tensor $(x, y, u, v)$, the conventional design is based on a 3D cost volume -- a 2D array $(x, y)$ augmented with a $uv$ channel, which is computationally efficient but often yields limited accuracy and overfitting.
To overcome this limitation, Yang~\etal~\cite{chJH:Yang:2019:VCN} proposed Volumetric Correspondence Networks (\textbf{VCN}), which are based on true 4D volumetric processing: constructing a proper 4D cost volume and processing with 4D convolution kernels.
For reducing the computational cost and memory of 4D processing, Yang~\etal~\cite{chJH:Yang:2019:VCN} used separable 4D convolutions, which approximate the 4D convolution operation with two 2D convolutions, reducing the complexity by $N^2$ (please refer the original paper for technical details).
Through proper 4D volumetric processing with computationally cheaper operations, the method further pushes both accuracy and practicality on widely used public benchmarks, improving generalization and demonstrating faster training convergence -- requiring $7$ times fewer training iterations than its competitors.

\begin{table}[t]
\caption{Quantitative comparison on public benchmarks: MPI Sintel \cite{chJH:Butler:2012:NOS} and KITTI \cite{chJH:Geiger:2012:AWR,chJH:Menze:2015:OSF}.}
\label{chJH:tab:genericCNN_eval}
{
\setlength\tabcolsep{12pt}
\scriptsize
\begin{tabularx}{\textwidth}{@{}XS[table-format=1.3]S[table-format=1.3]S[table-format=2.2,table-space-text-post={$\,\%$},table-align-text-post=true]S[table-format=2.2,table-space-text-post={$\,\%$},table-align-text-post=true]@{}}
\toprule
\multirow{2}{*}{Methods} & \multicolumn{2}{c}{MPI Sintel $^a$} & \multicolumn{2}{c@{}}{KITTI $^b$} \\\cmidrule(lr){2-3}\cmidrule(l){4-5}
	& {Clean} & {Final} & {2012} & {2015} \\
\midrule
FlowNetS \cite{chJH:Dosovitskiy:2015:FLO} & 6.158 & 7.218 & 37.05\,\% & {--} \\
FlowNetC \cite{chJH:Dosovitskiy:2015:FLO} & 6.081 & 7.883 & {--} & {--} \\
SPyNet \cite{chJH:Ranjan:2017:OFE} & 6.640 & 8.360 & 12.31\,\% & 35.07\,\% \\
FlowNet2 \cite{chJH:Ilg:2017:FN2} & 3.959 & 6.016 & 4.82\,\% & 10.41\,\% \\
PWC-Net \cite{chJH:Sun:2018:PCO} & 4.386 & 5.042 & 4.22\,\% & 9.60\,\% \\
LiteFlowNet \cite{chJH:Hui:2018:LFN} & 3.449 & 5.381 & 3.27\,\% & 9.38\,\% \\
IRR-PWC \cite{chJH:Hur:2019:IRR} & 3.844 & 4.579 & 3.21\,\% & 7.65\,\% \\
HD$^3$ \cite{chJH:Yin:2019:HDD} & 4.788 & 4.666 & \bfseries 2.26\,\% & 6.55\,\% \\
VCN \cite{chJH:Yang:2019:VCN} & \bfseries 2.808 & \bfseries 4.404 & {--} & \bfseries 6.30\,\% \\

\bottomrule
\end{tabularx}

$^a$ Evaluation metric: end point error (EPE).

$^b$ Evaluation metric: outlier rate (\ie~less than 3 pixel or 5\% error is considered an inlier)
}
\end{table}

{
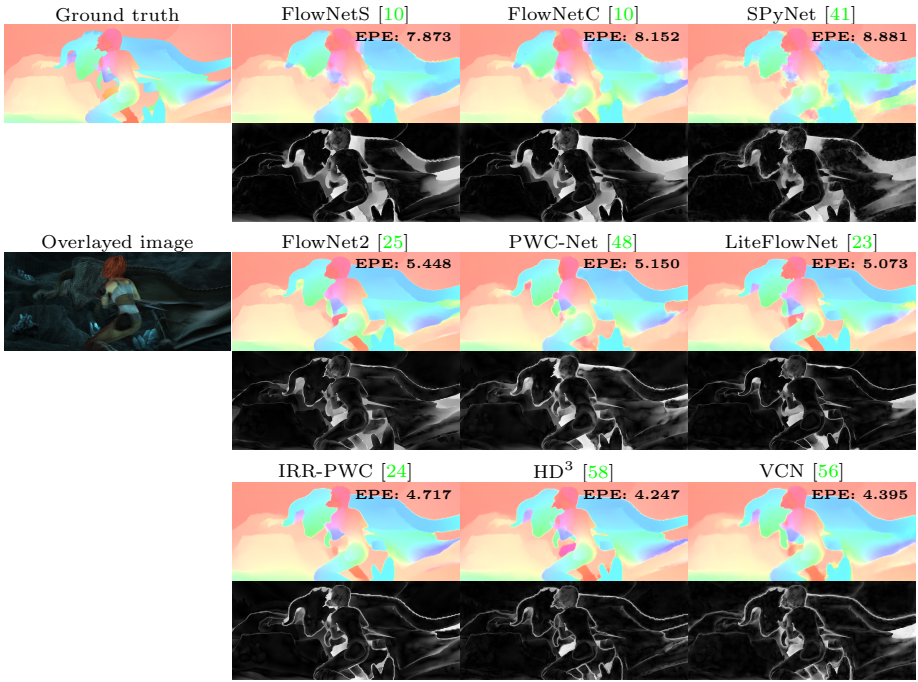
\begin{figure*}[t]
\centering
\scriptsize
\setlength\tabcolsep{0.3pt}
\renewcommand{\arraystretch}{0.2}
\begin{tabular}{>{\centering\arraybackslash}m{.248\textwidth} >{\centering\arraybackslash}m{.248\textwidth} >{\centering\arraybackslash}m{.248\textwidth} >{\centering\arraybackslash}m{.248\textwidth}}

Ground truth &
FlowNetS \cite{chJH:Dosovitskiy:2015:FLO} &
FlowNetC \cite{chJH:Dosovitskiy:2015:FLO} &
SPyNet \cite{chJH:Ranjan:2017:OFE} \\
\includegraphics[width=\linewidth]{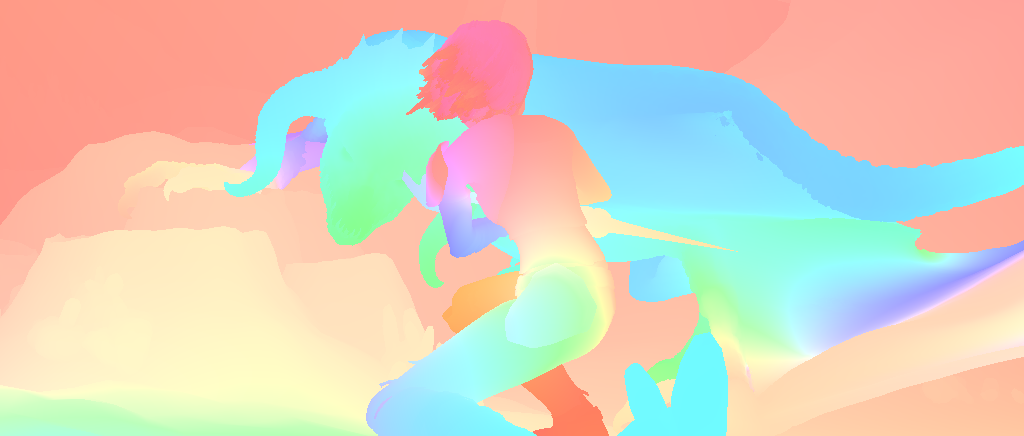} &
\begin{tikzpicture}
\node[inner sep=0] (img) {\includegraphics[width=\linewidth]{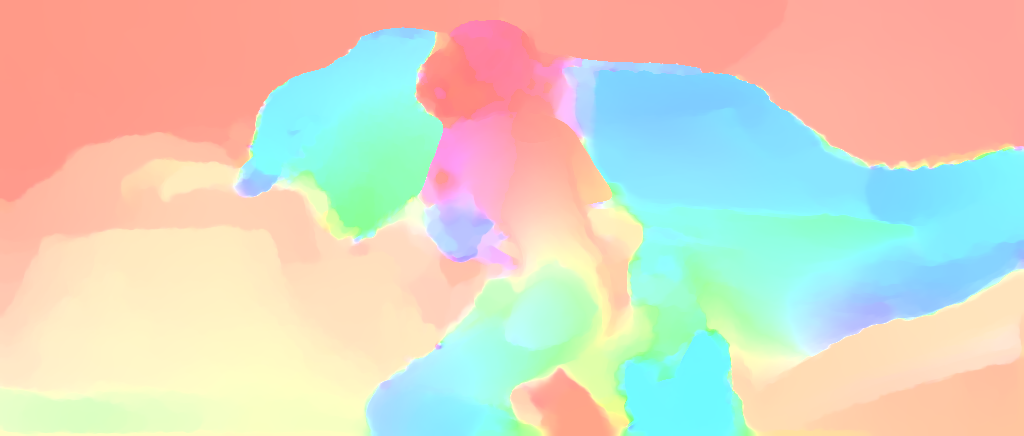}};
\node[anchor=north east] at (img.north east){\tiny \color{black} \textbf{EPE: 7.873}};
\end{tikzpicture} & 	
\begin{tikzpicture}
\node[inner sep=0] (img) {\includegraphics[width=\linewidth]{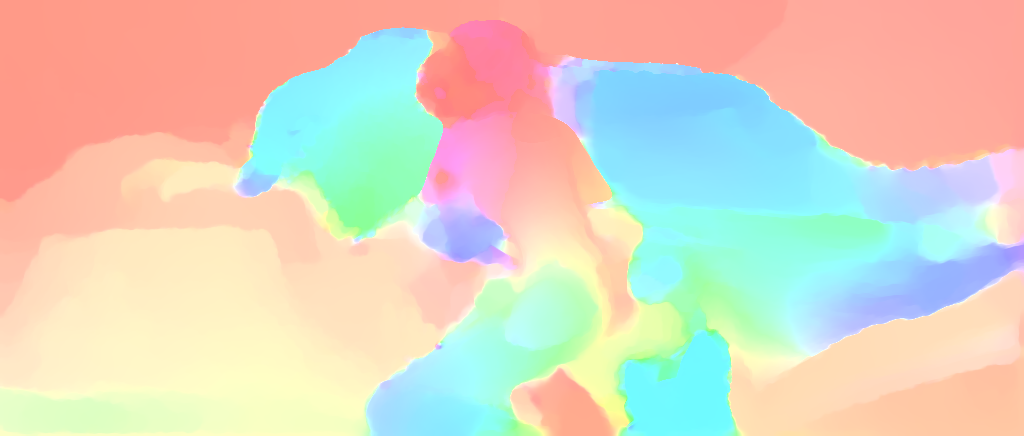}};
\node[anchor=north east] at (img.north east){\tiny \color{black} \textbf{EPE: 8.152}};
\end{tikzpicture} & 
\begin{tikzpicture}
\node[inner sep=0] (img) {\includegraphics[width=\linewidth]{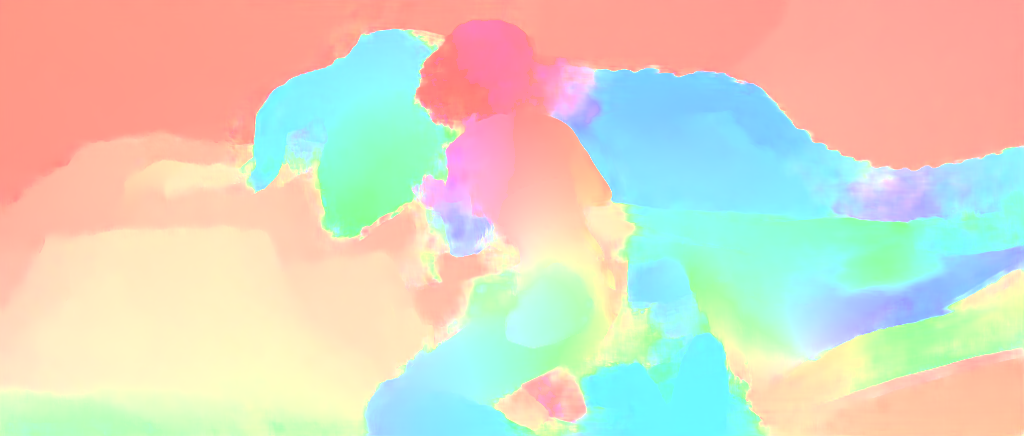}};
\node[anchor=north east] at (img.north east){\tiny \color{black} \textbf{EPE: 8.881}};
\end{tikzpicture} \\ 
&
\includegraphics[width=\linewidth]{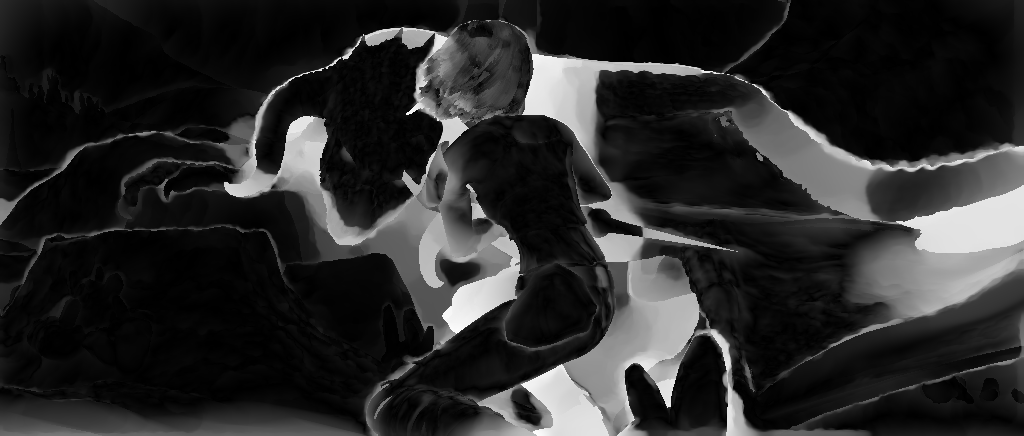} & 	
\includegraphics[width=\linewidth]{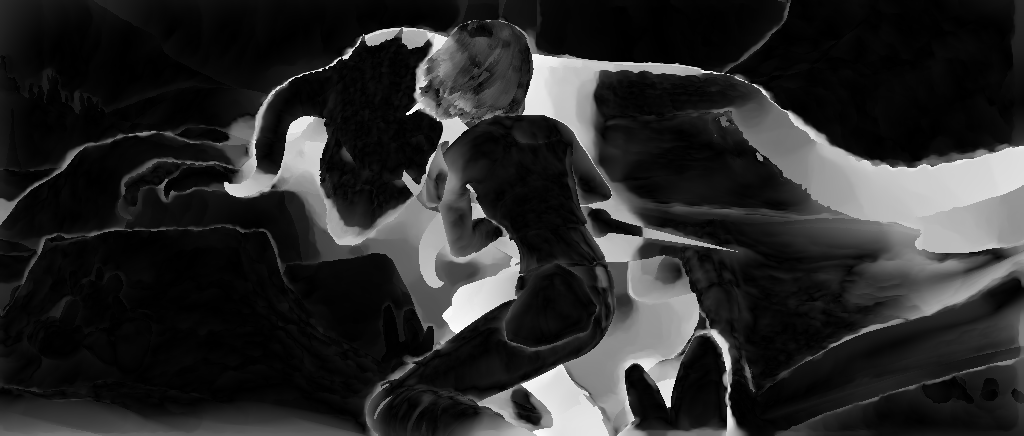} & 
\includegraphics[width=\linewidth]{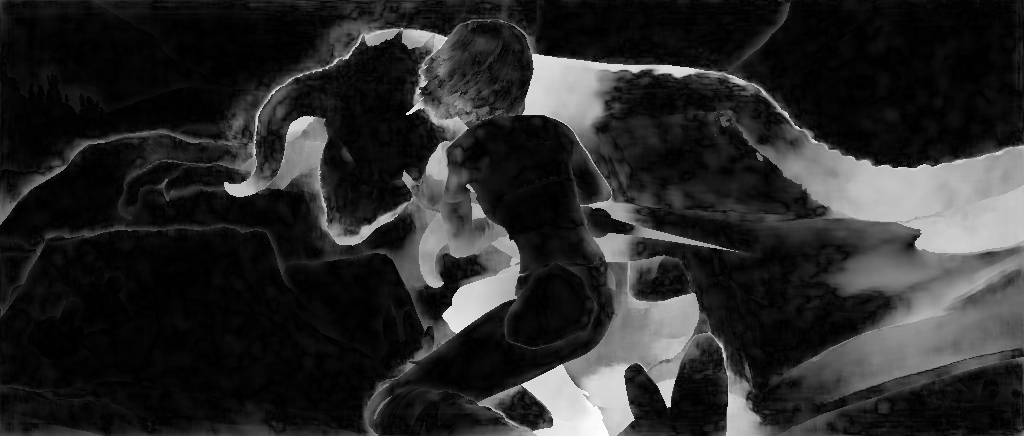} \\ \; \vspace{2pt} \\

Overlayed image &
FlowNet2 \cite{chJH:Ilg:2017:FN2} &
PWC-Net \cite{chJH:Sun:2018:PCO}  &
LiteFlowNet \cite{chJH:Hui:2018:LFN} \\
\includegraphics[width=\linewidth]{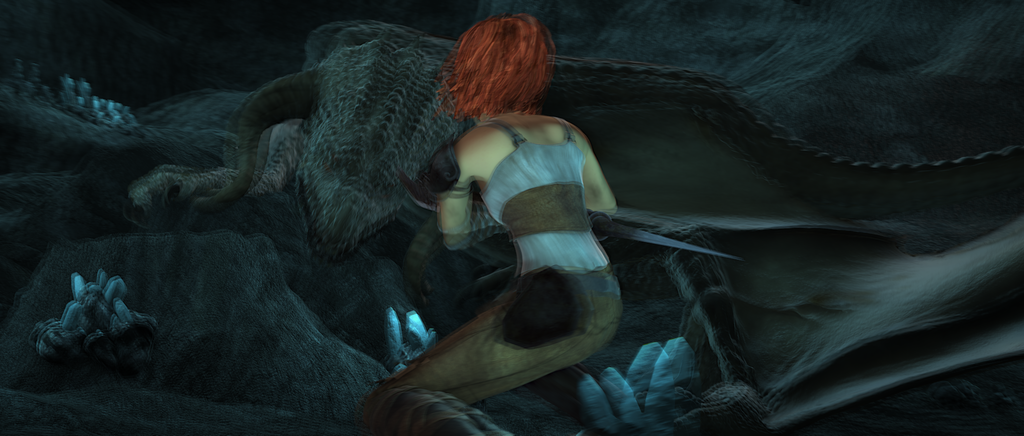} &
\begin{tikzpicture}
\node[inner sep=0] (img) {\includegraphics[width=\linewidth]{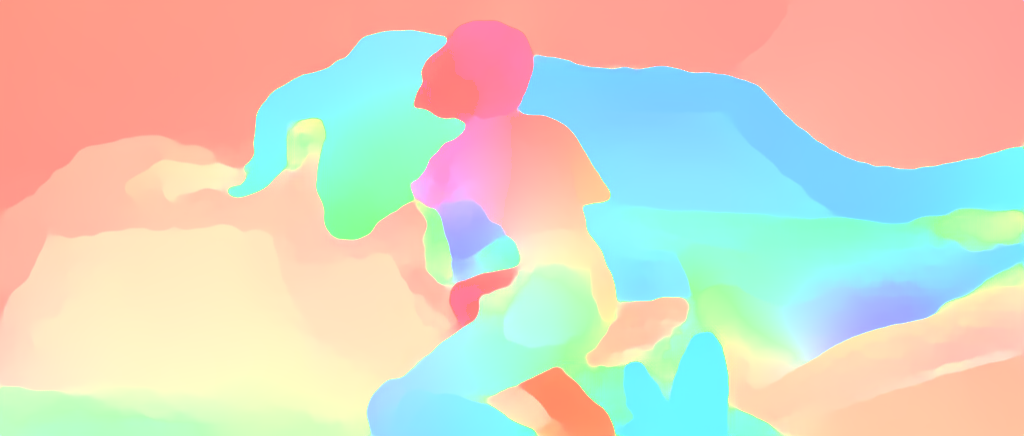}};
\node[anchor=north east] at (img.north east){\tiny \color{black} \textbf{EPE: 5.448}};
\end{tikzpicture} & 
\begin{tikzpicture}
\node[inner sep=0] (img) {\includegraphics[width=\linewidth]{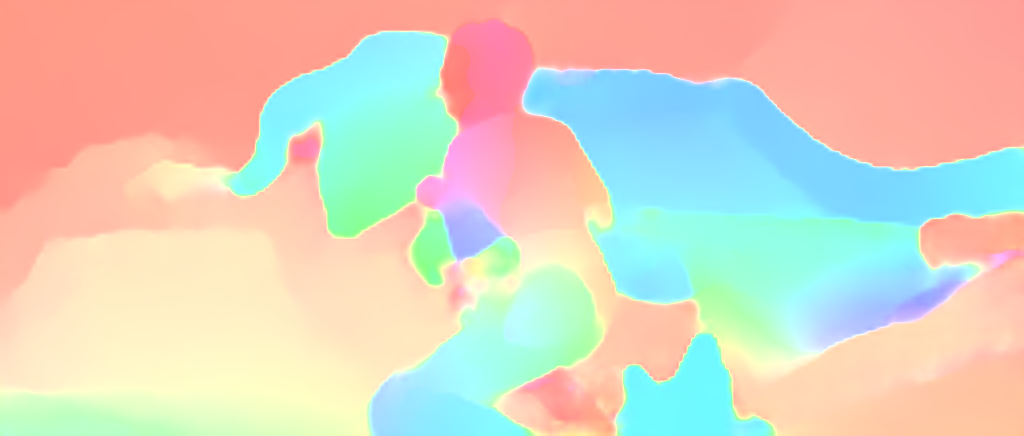}};
\node[anchor=north east] at (img.north east){\tiny \color{black} \textbf{EPE: 5.150}};
\end{tikzpicture} & 
\begin{tikzpicture}
\node[inner sep=0] (img) {\includegraphics[width=\linewidth]{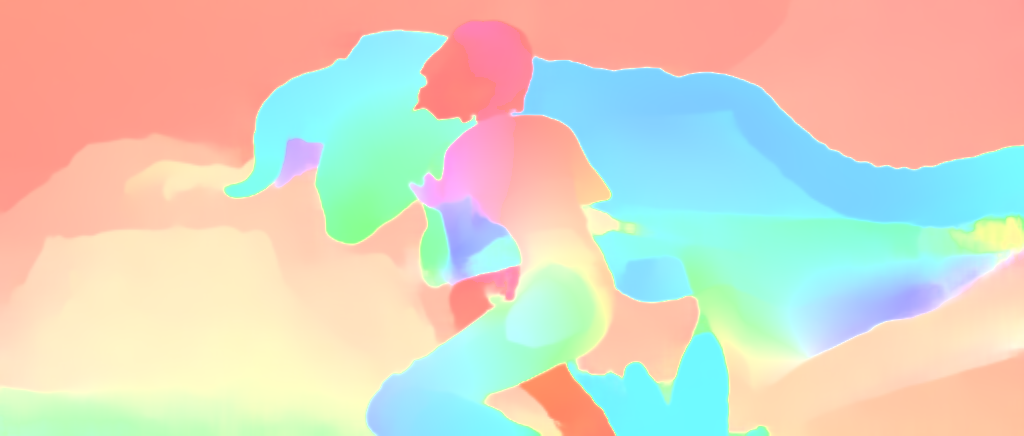}};
\node[anchor=north east] at (img.north east){\tiny \color{black} \textbf{EPE: 5.073}};
\end{tikzpicture} \\
&
\includegraphics[width=\linewidth]{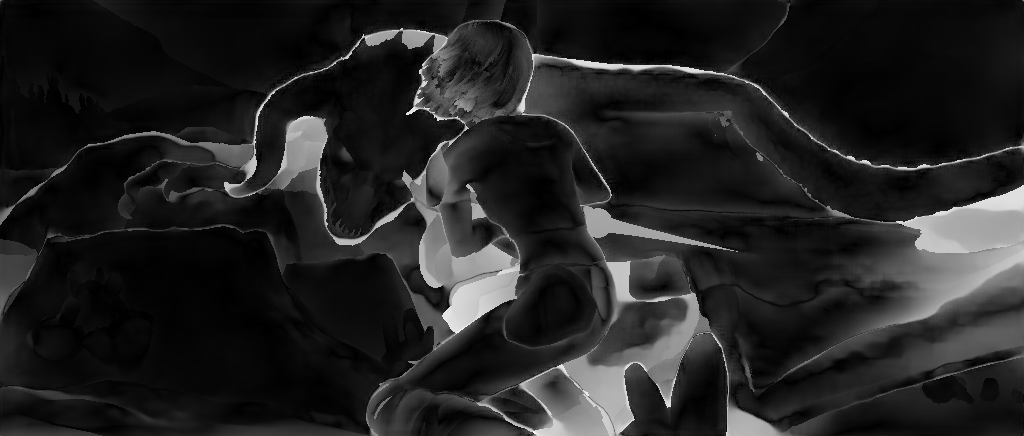} &
\includegraphics[width=\linewidth]{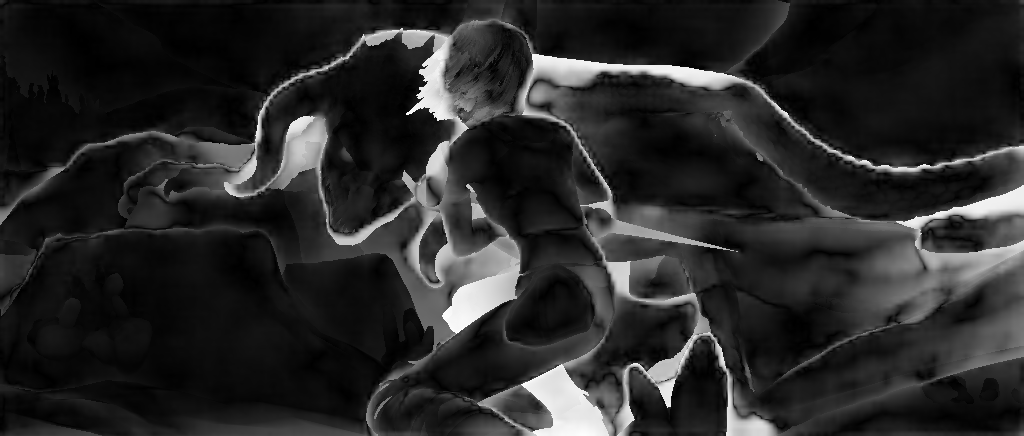} & 
\includegraphics[width=\linewidth]{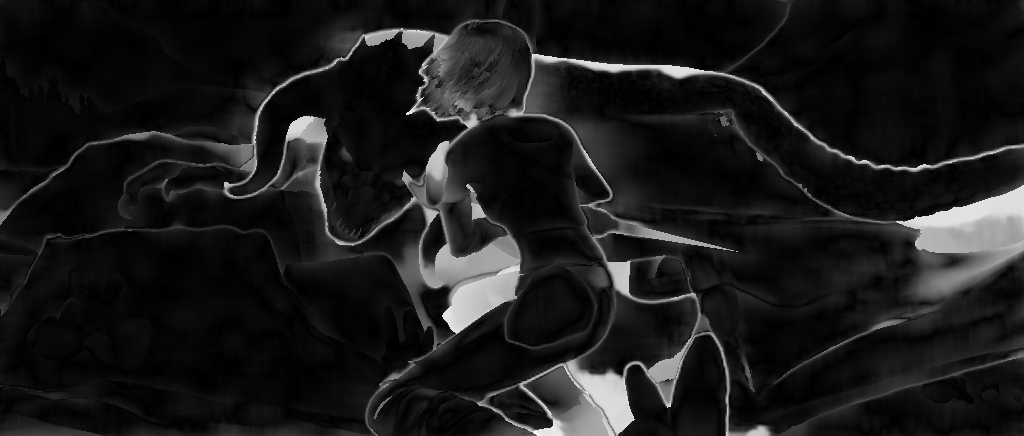} \\ \; \vspace{2pt} \\

&
IRR-PWC \cite{chJH:Hur:2019:IRR} &
HD$^3$ \cite{chJH:Yin:2019:HDD}  &
VCN \cite{chJH:Yang:2019:VCN} \\
 &
\begin{tikzpicture}
\node[inner sep=0] (img) {\includegraphics[width=\linewidth]{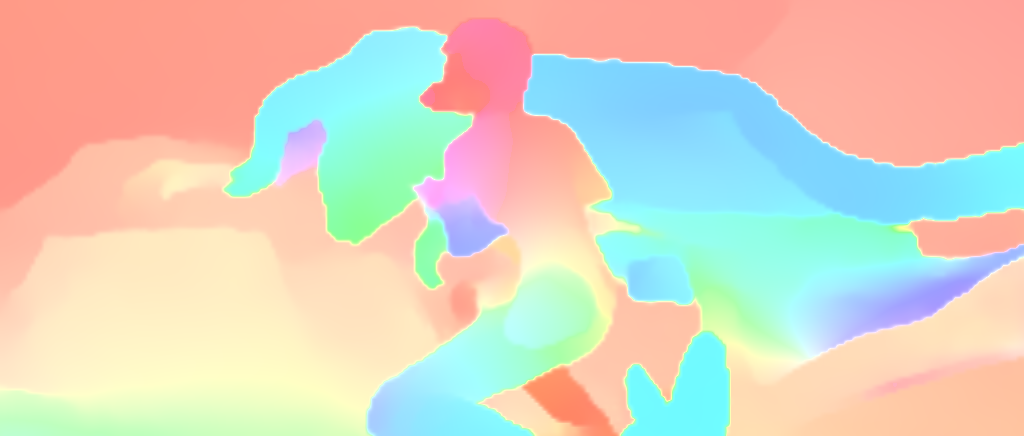}};
\node[anchor=north east] at (img.north east){\tiny \color{black} \textbf{EPE: 4.717}};
\end{tikzpicture} & 
\begin{tikzpicture}
\node[inner sep=0] (img) {\includegraphics[width=\linewidth]{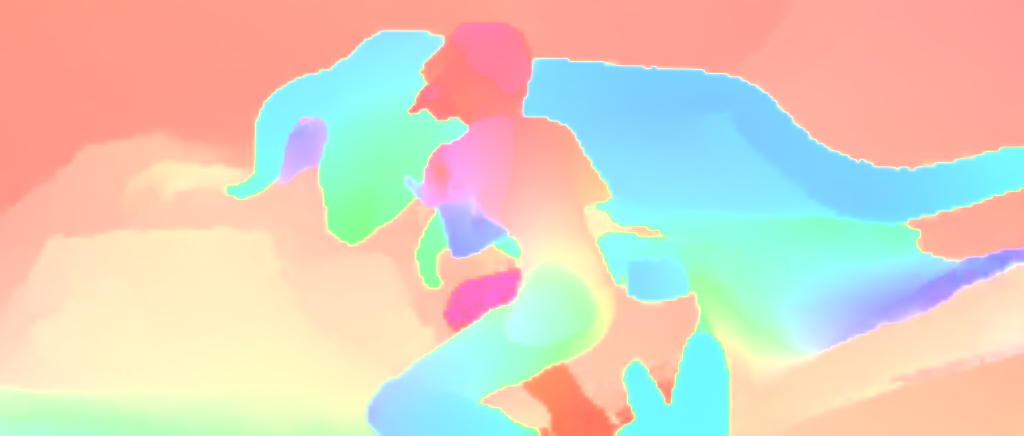}};
\node[anchor=north east] at (img.north east){\tiny \color{black} \textbf{EPE: 4.247}};
\end{tikzpicture} & 
\begin{tikzpicture}
\node[inner sep=0] (img) {\includegraphics[width=\linewidth]{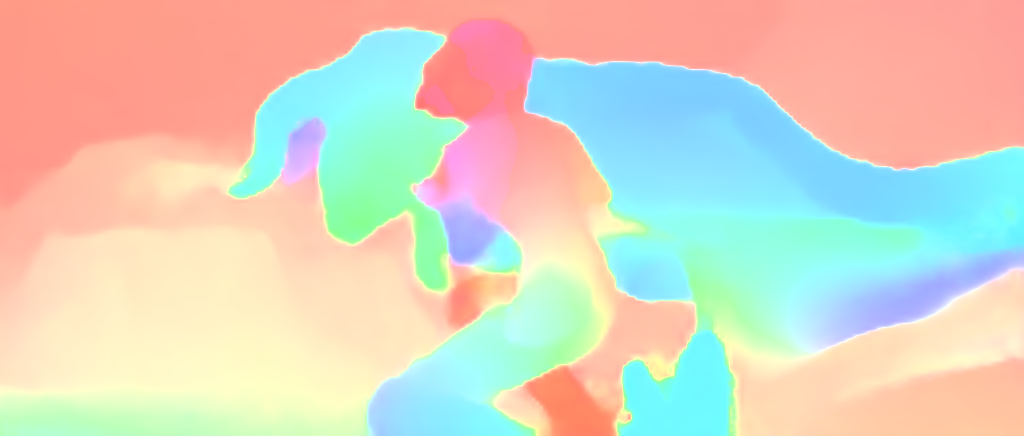}};
\node[anchor=north east] at (img.north east){\tiny \color{black} \textbf{EPE: 4.395}};
\end{tikzpicture} \\
&
\includegraphics[width=\linewidth]{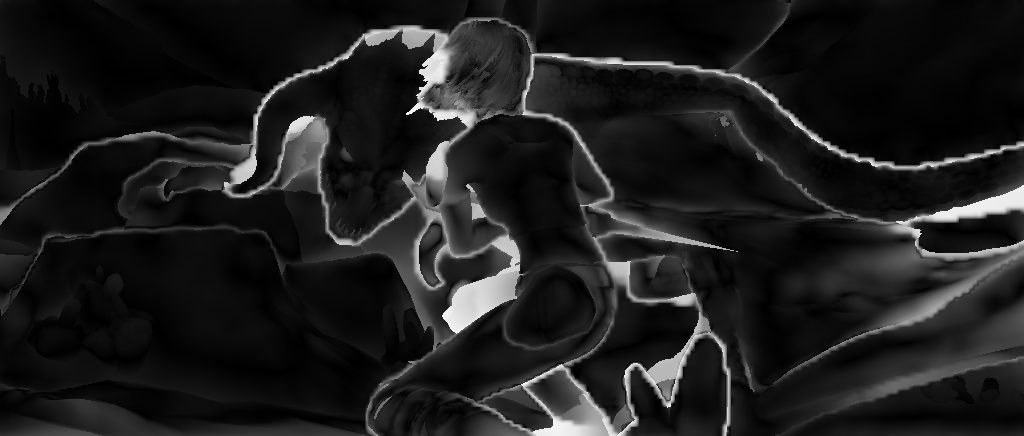} &
\includegraphics[width=\linewidth]{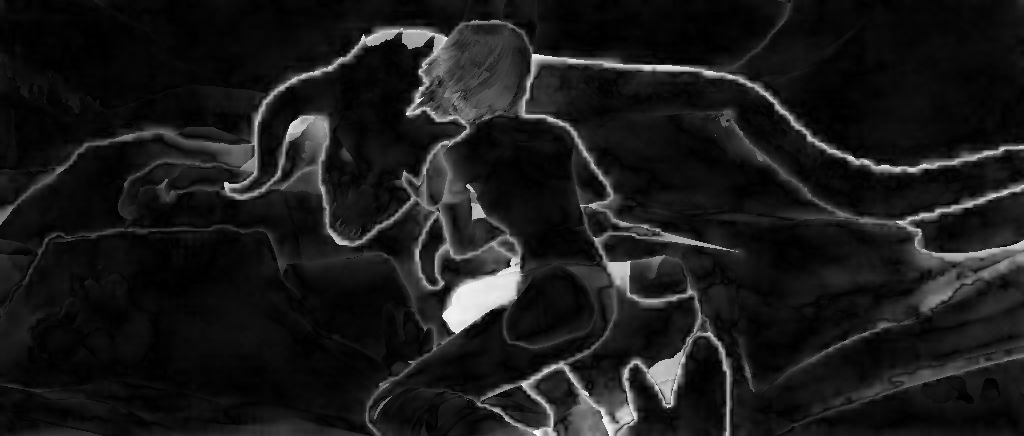} & 
\includegraphics[width=\linewidth]{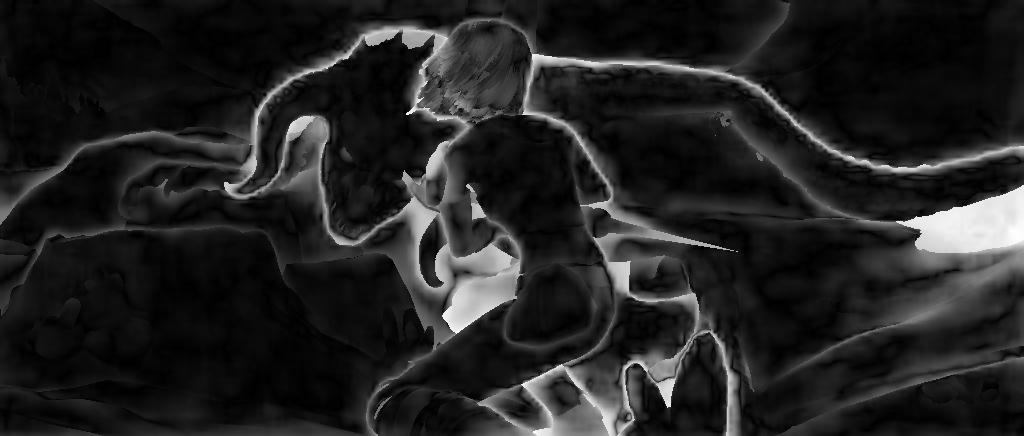} \\

\end{tabular}
\caption {\textbf{Qualitative comparison of end-to-end architectures}: Example from Sintel Final Test \cite{chJH:Butler:2012:NOS}. The first column shows the ground-truth flow and the overlayed input images. In the further columns, we show the color-coded flow visualization of each method, overlayed with the end point error (EPE) and their error maps (the brighter a pixel, the higher its error).}
\label{chJH:fig:qualitative_sintel}
\end{figure*}
}

Table~\ref{chJH:tab:genericCNN} summarizes the main differences in technical design of the various end-to-end optical flow architectures discussed above.
Starting from FlowNetS \cite{chJH:Dosovitskiy:2015:FLO}, the methods are listed in chronological order. 
We omit the IRR scheme as it can be applied on top of several backbone architectures.
Table~\ref{chJH:tab:genericCNN_eval} compares the quantitative results of each method on the MPI Sintel \cite{chJH:Butler:2012:NOS} and KITTI benchmarks \cite{chJH:Geiger:2012:AWR,chJH:Menze:2015:OSF}.
Each method is pre-trained on synthetic datasets first and then fine-tuned on each benchmark.
Looking at the two tables, we can gain some first insights into which design choices lead to the observed accuracy improvements.
First, having a pyramid structure by adopting a ``coarse-to-fine'' strategy makes networks more compact and improves the flow estimation accuracy (\eg,~from FlowNet \cite{chJH:Dosovitskiy:2015:FLO} to SPyNet \cite{chJH:Ranjan:2017:OFE}, PWC-Net \cite{chJH:Sun:2018:PCO}, and LiteFlowNet \cite{chJH:Hui:2018:LFN}).
Second, stacking networks can also improve the flow accuracy while linearly increasing the number of parameters (\eg,~from FlowNet \cite{chJH:Dosovitskiy:2015:FLO} to FlowNet2 \cite{chJH:Ilg:2017:FN2}).
Third, constructing a cost volume by calculating a patch-wise correlation between two feature maps has become a standard approach and is more beneficial than not using it (\eg,~FlowNetS vs.~FlowNetC, according to a study from \cite{chJH:Ilg:2017:FN2}).
Fourth, even if based on similar conceptual designs, subtle design differences or additional modules can further lead to accuracy improvements (\eg,~LiteFlowNet \cite{chJH:Hui:2018:LFN} vs.~PWC-Net \cite{chJH:Sun:2018:PCO}).
Fifth, the iterative residual refinement scheme IRR \cite{chJH:Hur:2019:IRR} can further boost the accuracy of existing backbone architectures (\eg,~from PWC-Net \cite{chJH:Sun:2018:PCO} to IRR-PWC \cite{chJH:Hur:2019:IRR}). 
Lastly, investigating better fundamental designs such as the output representation (\eg, the match density \cite{chJH:Yin:2019:HDD}) or the cost volume representation (\eg, 4D cost volume \cite{chJH:Yang:2019:VCN}) can lead to further improvement, sometimes quite significantly so.

Fig.~\ref{chJH:fig:qualitative_sintel} shows a qualitative comparison of each method on an example from the Sintel Final Test set \cite{chJH:Butler:2012:NOS}.
The optical flow visualizations and the error maps demonstrate how significantly end-to-end methods have been improved over the past few years, especially near motion boundaries and in non-textured areas.

%% file: chJH_sec2.tex
\section{Approaches with Alternative Learning Paradigms}
\label{chJH_sec:2}

Aside from the question of how to design deep network architectures for optical flow estimation, another problem dimension has grown into prominence recently -- how to train such CNNs for optical flow especially in the context of the limited quantities of ground-truth data available in practice.
Most (early) CNN approaches are based on standard supervised learning and directly train the network on labeled data.
However, real-world labeled data is available only in comparatively small quantities and often constrained to certain settings, which turns out to have the limitation that the accuracy can drop significantly on unseen data.
To overcome this, a number of alternative approaches based on unsupervised or semi-supervised learning have been proposed to lighten the necessity of and reliance on large amounts of labeled data.
In this section, we review and categorize CNN approaches in terms of their underlying learning paradigm: supervised learning, unsupervised or self-supervised learning, and finally semi-supervised learning.

\subsection{Challenges of supervised learning}
\label{chJH_subsec:2_1}
Based on the end-to-end trainability of CNNs, the most straightforward way to train CNNs for optical flow estimation is in a supervised fashion using a labeled dataset.
In the supervised learning setting -- but not only there -- the dataset plays an important role, and details such as the size and design of the dataset, the type of loss function, and training schedules become critical factors in achieving high accuracy.

Approaches that are based on CNNs as feature extractor \cite{chJH:Bai:2016:ESI,chJH:Bailer:2017:CPM,chJH:Gadot:2016:PBB,chJH:Gueney:2016:DDF}, as already discussed above, collect positive matching samples and negative non-matching samples as a training set and train the CNNs by applying a loss function at the final output of the network.
Different types of loss functions has been investigated to obtain discriminative features that are invariant to common appearance and illumination changes (please refer to Sec.~\ref{chJH_subsec:1_2} for further details).
When training CNNs in general, having a large labeled dataset is crucial to avoid overfitting on the training dataset and enable the network to generalize to unseen data.
As the networks tend to be comparatively lean and do not have to (and in fact cannot) learn something about plausible motions, but rather only classify when patches match in terms of their appearance, the issue of overfitting is less prominent than in end-to-end regression approaches.

For training end-to-end optical flow architectures in a supervised fashion, on the other hand, we need to have a training dataset with many temporally consecutive image pairs with dense ground-truth flow, representing the range of possible optical flow fields. 
The entire flow map with per-pixel labels is used to train the network by minimizing the per-pixel Euclidean distance between the ground truth flow and the output from the network.
However, collecting such a dataset with real-world images has been challenging due to the difficulty of measuring the true motion for all pixels \cite{chJH:Baker:2011:DBE}. 
Establishing synthetic datasets instead is a viable alternative (\eg,~the FlyingChairs \cite{chJH:Dosovitskiy:2015:FLO}, Sintel \cite{chJH:Butler:2012:NOS}, and FlyingThings3D \cite{chJH:Mayer:2016:ALD} datasets), as it is much easier to generate a large amount of synthesized images with accurate ground-truth flow.

Yet, using a synthetic dataset for training flow networks still does not completely solve the issue of dataset suitability.
The generalization to an unseen setting remains a challenge.
According to the empirical studies of \cite{chJH:Ilg:2017:FN2} and \cite{chJH:Sun:2018:PCO}, the flow accuracy significantly depends on the dataset used for training and on how close the test-time domain is to the training domain.
Consequently, overfitting on the training dataset domain is a problem.
As a solution, FlowNet2 \cite{chJH:Ilg:2017:FN2} is accompanied with a training dataset schedule that leads to a better local parameter optimum so that the trained networks can perform reasonably on unseen data:
pre-training on synthetic datasets before fine-tuning on the target domain dataset in the end (please refer to Sec.~\ref{chJH_subsec:1_3} for further details).
Both FlowNet2 \cite{chJH:Ilg:2017:FN2} and PWC-Net \cite{chJH:Sun:2018:PCO} empirically demonstrated that training networks with this schedule allows for better generalization to an unseen target domain.
In fact, pre-training on a synthetic dataset followed by fine-tuning on the target domain yields much better accuracy than directly training on the target domain, even on the target domain itself.

All regression architectures mentioned above have multi-scale intermediate optical flow outputs along the decoder (\eg,~FlowNet~\cite{chJH:Dosovitskiy:2015:FLO} and FlowNet2~\cite{chJH:Ilg:2017:FN2}) or at each pyramid level (\eg,~PWC-Net~\cite{chJH:Sun:2018:PCO}, SPyNet~\cite{chJH:Ranjan:2017:OFE}, and LiteFlowNet~\cite{chJH:Hui:2018:LFN}). For all intermediate outputs, an $L_2$ loss between the output and the downscaled ground truth is applied per pixel so that the network learns to estimate optical flow in a coarse-to-fine manner and achieves better accuracy at the final output resolution.
The final training loss becomes the weighted sum of all intermediate losses.

\subsection{Unsupervised or self-supervised learning}
\label{chJH_subsec:2_2}

While synthetic datasets enable training CNNs with a large amount of labeled data, the networks only trained on synthetic datasets perform relatively poorly on real-world datasets due to the domain mismatch between the training domain and the target domain.
As just discussed, supervised approaches thus require fine-tuning on the target domain for better accuracy.
However, this can be problematic if there is no ground truth optical flow  available for the target domain.
To resolve this issue, unsupervised learning approaches have been proposed to directly train CNNs on the target domain without having access to any ground truth flow.
Such methods are also called self-supervised, as the supervisory signal comes from the input images themselves.
In this section, we will overview existing unsupervised or self-supervised learning methods and discuss how they have progressed to achieve results that are competitive with many supervised methods.

Ahmadi and Patras \cite{chJH:Ahmadi:2016:UCN} pioneered unsupervised learning-based optical flow using CNNs.
Inspired by the classical Horn and Schunck \cite{chJH:Horn:1981:DOF} method, Ahmadi and Patras used the classical optical flow constraint equation as a loss function for training the network.
By minimizing this unsupervised loss function, the network learns to predict optical flow fields that satisfy the optical flow constraint equation on the input images, \ie,~the brightness constancy assumption.
\cite{chJH:Ahmadi:2016:UCN} further combines this with classical coarse-to-fine estimation so that the flow field improves through multi-scale estimation.
By demonstrating that the flow accuracy is close to the best supervised method at the time, \ie~FlowNet \cite{chJH:Dosovitskiy:2015:FLO}, Ahmadi and Patras \cite{chJH:Ahmadi:2016:UCN} suggest that unsupervised learning of networks for optical flow estimation is possible and can overcome some of the limitations of supervised learning approaches.

Concurrently, Yu~\etal~\cite{chJH:Yu:2016:BBU} and Ren~\etal~\cite{chJH:Ren:2017:UDL} proposed to use a proxy unsupervised loss that is inspired by a standard MRF formulation. 
Following classical concepts, the proposed unsupervised proxy loss consists of a data term and a smoothness term as in Eq.~\eqref{chJH:eq:Horn_Schunck_mrf}.
The data term directly minimizes the intensity difference between the first image and the warped second image from estimated optical flow, and the smoothness term penalizes flow differences between neighboring pixels.
Both methods demonstrate that directly training on a target domain (\eg,~the KITTI datasets \cite{chJH:Geiger:2012:AWR}) in an unsupervised manner performs competitive to or sometimes even outperforms the same network that is trained on a different domain (\eg,~the FlyingChairs dataset \cite{chJH:Dosovitskiy:2015:FLO}) in a supervised manner.
This observation suggests that unsupervised learning approaches can be a viable alternative to supervised learning, if labeled data for training is not available in the target domain.

In a follow-up work, Zhu~\etal~\cite{chJH:Zhu:2017:DDF} showed that the backbone network can be improved by using a dense connectivity.
They built on DenseNet \cite{Huang:2017:DCC}, which uses dense connections with skip connections between all convolutional layers to improve the accuracy over the previous state of the art for image classification.
Inspired by DenseNet, Zhu~\etal~\cite{chJH:Zhu:2017:DDF} adopted the such dense connections in an hourglass-shaped architecture by using dense blocks before every downsampling and upsampling step; each dense block has four convolutional layers with dense skip connections between each other.
\cite{chJH:Zhu:2017:DDF} improves the flow accuracy by more than 10\% on public benchmark datasets over \cite{chJH:Yu:2016:BBU} on average, which uses FlowNet \cite{chJH:Dosovitskiy:2015:FLO} as a backbone network, indicating the importance of choosing the right backbone network in the unsupervised learning setting as well.

Zhu~\etal~\cite{chJH:Zhu:2017:GOF} also proposed a different direction of unsupervised learning, combining an unsupervised proxy loss and a guided supervision loss using proxy ground truth obtained from an off-the-shelf classical energy-based method.
As in \cite{chJH:Ren:2017:UDL,chJH:Yu:2016:BBU}, the unsupervised proxy loss makes the network learn to estimate optical flow to satisfy the brightness constancy assumption while the guided loss helps the network perform close to off-the-shelf classical energy-based method.
In the circumstance that learning with the unsupervised proxy loss is outperformed by the classical energy-based method, the guided loss can help and even achieve better accuracy than either of the two losses alone.

Unsupervised or self-supervised learning of optical flow relies on minimizing a proxy loss rather than estimating optical flow close to some ground truth.
Thus, designing a faithful proxy loss is critical to its success.
Meister~\etal~\cite{chJH:Meister:2018:ULO} proposed a proxy loss function that additionally considers occlusions, demonstrates better accuracy than previous unsupervised methods, and outperforms the supervised backbone network (\ie,~FlowNet \cite{chJH:Dosovitskiy:2015:FLO}).
Further, bi-directional flow is estimated from the same network by only switching the order of input images and occlusions are detected using a bi-directional consistency check.
The proxy loss is applied only to non-occluded regions as the brightness constancy assumption does not hold for occluded pixels.
In addition, Meister~\etal~\cite{chJH:Meister:2018:ULO} suggested to use a higher-order smoothness term and a ternary census loss \cite{Stein:2004:ECO,Zabih:1994:NPL} to obtain a data term that is robust to brightness changes.
This advanced proxy loss significantly improves the accuracy by halving the error compared to previous unsupervised learning approaches.
\cite{chJH:Meister:2018:ULO} resulting in better accuracy than supervised approaches pre-trained on synthetic data alone (assuming the same backbone), which suggests that directly training on the target domain in an unsupervised manner can be a good alternative to supervised pre-training with synthetic data.

Wang~\etal~\cite{chJH:Wang:2018:OAU} also introduced an advanced proxy loss that takes occlusion into account and is applied only to non-occluded regions.
Similar to \cite{chJH:Meister:2018:ULO}, Wang~\etal~\cite{chJH:Wang:2018:OAU} estimate bi-directional optical flow and then obtain an occlusion mask for the forward motion by directly calculating disocclusion from the backward flow.
They exploit the fact that occlusion from the forward motion is the inverse of disocclusion from the backward motion.
Disocclusions can be obtained by forward-warping the given flow and detecting the holes to which no pixels have been mapped.
In addition to occlusion handling, their approach contains other innovations such as a modified architecture and pre-processing.
According to their ablation study, the accuracy is improved overall by 25\% on public benchmark datasets compared to the unsupervised approach of Yu~\etal~\cite{chJH:Yu:2016:BBU}.
In addition, the method demonstrates good occlusion estimation results, close to those of classical energy-based approaches. 

Janai~\etal~\cite{chJH:Janai:2018:ULM} extended unsupervised learning of optical flow to a multi-frame setting, taking in three consecutive frames and jointly estimating an occlusion map.
Based on the PWC-Net \cite{chJH:Sun:2018:PCO} architecture, they estimate bi-directional flow from the reference frame and occlusion maps for both directions as well.
After the cost volume of PWC-Net, Janai~\etal~use three different decoders:
\emph{(i)} a future frame decoder that estimates flow from the reference frame to the future frame, \emph{(ii)} a past flow decoder, and \emph{(iii)} an occlusion decoder.
A basic unsupervised loss consisting of photometric and smoothness terms is applied only on non-occluded regions for estimating flow, and a constant velocity constraint is also used, which encourages the magnitude of forward flow and backward flow to be similar but going in opposite directions.
Their experimental results demonstrate the benefits of using multiple frames, outperforming all two-frame based methods.
Furthermore, the accuracy of occlusion estimation is competitive with classical energy-based methods.

Liu~\etal~\cite{chJH:Liu:2019:DLO,chJH:Liu:2019:SFS} demonstrated another direction for unsupervised (or self-supervised) learning by using a data distillation framework with student-teacher networks.
Their two methods, DDFlow \cite{chJH:Liu:2019:DLO} and its extension SelFlow \cite{chJH:Liu:2019:SFS}, distill reliable predictions from a teacher network, which is trained in an unsupervised manner \cite{chJH:Meister:2018:ULO}, and use them as pseudo ground truth for training the student network, which is used at inference time.
The accuracy of this framework depends on how to best distill the knowledge for the student network.
For better accuracy especially in occluded regions, the two methods focus on how to provide more reliable labels for occluded pixels to the student network.
DDFlow \cite{chJH:Liu:2019:DLO} proposes to randomly crop the predicted flow map from the teacher network as well as the input images.
Then in the cropped images, some of the non-occluded pixels near the image boundaries become out-of-bounds pixels (\ie, occluded pixels), and its reliably predicted optical flow from the non-occluded pixels in the teacher network can work as reliable pseudo ground truth for occluded pixels in the student network.
In the experiments, DDFlow \cite{chJH:Liu:2019:DLO} showed data distillation to significantly improve the accuracy on average up to $34.7\%$ on public benchmark datasets, achieving the best accuracy among existing unsupervised learning-based approaches.

SelFlow \cite{chJH:Liu:2019:SFS} suggests a better data distillation strategy by exploiting superpixel knowledge and hallucinating occlusions in non-occluded regions.
Given the prediction from the teacher network, SelFlow \cite{chJH:Liu:2019:SFS} superpixelizes the target frame and perturbs random superpixels by injecting random noise as if non-occluded pixels in the target images were occluded by randomly looking superpixels.
Then likewise, those non-occluded pixels with reliable predictions from the teacher network become occluded pixels when training the student network, guiding to estimate reliable optical flow in occluded areas.
In addition, SelFlow \cite{chJH:Liu:2019:SFS} further demonstrates multi-frame extensions using 3 frames as input for improving the accuracy by exploiting temporal coherence.
Evaluating on public benchmark datasets, SelFlow \cite{chJH:Liu:2019:SFS} further improves the accuracy over DDFlow \cite{chJH:Liu:2019:DLO}, demonstrating the importance of having a better data distillation strategy and suggesting a promising direction for self-supervised learning.

\subsection{Semi-supervised learning}
\label{chJH_subsec:2_3}
Complementary to supervised and unsupervised learning methods, semi-supervised learning approaches have been also proposed recently.
Lai~\etal~\cite{chJH:Lai:2017:SLO} utilized Generative Adversarial Networks (GANs) \cite{chJH:Goodfellow:2014:GAN} and proposed an adversarial loss that captures the structural pattern of the flow warp error, allowing to train a network in a semi-supervised way. 
First, a generator network produces optical flow from the two given input images.
Next, the flow warp error map is obtained by calculating the image intensity difference between the first image and the warped second image using the flow output.
Then, a discriminator network tries to distinguish whether the warp error map is created by the generator or is the ground truth.
The generator aims to fool the discriminator network by producing optical flow whose warp error patterns look close to those of the ground truth.
Meanwhile, the discriminator keeps trying to correctly distinguish whether the flow warp error pattern is from the generated flow or the ground truth flow, challenging the generator. 
To train the networks, a combination of labeled and unlabeled data has been used, equally distributed in each mini-batch.
For labeled data in each mini-batch, the standard $L_2$ loss is applied to the output of the generator to ensure closeness of the flow estimate to the ground truth.
The adversarial loss is applied to the output of the discriminator to both labeled and unlabeled data.
The experiments demonstrate benefits over purely supervised and purely unsupervised methods: the results are more accurate than when training with a synthetic dataset only in a supervised way and they also outperform training with unlabeled real data in the target domain only in an unsupervised way.

Yang~\etal~\cite{chJH:Yang:2018:CPN} proposed another semi-supervised approach by learning a conditional prior for predicting optical flow.
They posit that current learning-based approaches to optical flow do not rely on any explicit regularizer (which refers to any prior, model, or assumption that adds any restrictions to the solution space), which results in a risk of overfitting on the training domain, relating to the domain mismatch problem regarding the testing domain.
To address the issue, they propose a network that contains prior information of possible optical flows that an input image can give rise to and then use the network as a regularizer for training a standard off-the-shelf optical flow network.
They first train the conditional prior network in a supervised manner to learn prior knowledge on the possible optical flows of an input image, and then train FlowNet \cite{chJH:Dosovitskiy:2015:FLO} in an unsupervised manner with a regularization loss from the trained conditional prior network.
The experiments demonstrate that the conditional prior network enables the same network trained on the same dataset \emph{(i)} to outperform typical unsupervised training and \emph{(ii)} to give results that are competitive with the usual supervised training, yet showing better generalization across different dataset domains.
This observation suggests that semi-supervised learning can benefit domain generalization without labeled data by leveraging the available ground truth from another domain.

%% file: chJH_sec3.tex

\section{Multi-frame Optical Flow Estimation}
\label{chJH_sec:3}
In the literature of classical optical flow methods, utilizing multiple frames has a long history (\eg, \cite{chJH:Nagel:1990:EOS}).
When additional temporally consecutive frames are available, different kinds of assumptions and strategies can be exploited.
One basic and straightforward way is to utilize the temporal coherence assumption that optical flow smoothly changes over time \cite{chJH:Black:1991:RDM,chJH:Janai:2017:SFE,chJH:Kennedy:2015:OFG,chJH:Volz:2011:MTC,chJH:Werlberger:2009:AHO}. 
This property is sometimes also referred to as constant velocity or acceleration assumption.
Another way is to parameterize and model the trajectories of motion, which allows to exploit higher-level motion information instead of simply enforcing temporal smoothness on optical flow \cite{chJH:Chaudhury:1995:ATC,chJH:Garg:2013:AVA,chJH:Ricco:2012:DLM} in 2D.
Recently, there has been initial work on adopting these proven ideas in the context of deep learning to improve the flow accuracy.

Ren~\etal~\cite{chJH:Ren:2019:FAM} proposed a multi-frame optical flow network by extending the two-frame, state-of-the-art PWC-Net \cite{chJH:Sun:2018:PCO}.
Given three temporally consecutive frames, $I_{t-1}$, $I_{t}$, and $I_{t+1}$, the proposed method fuses the two optical flows from $I_{t-1}$ to $I_{t}$ and from $I_{t}$ to $I_{t+1}$ to exploit the temporal coherence between the three frames.
Each optical flow is obtained using PWC-Net.
In order to fuse the two optical flows, the method also estimates the flow from $I_{t}$ to $I_{t-1}$ to backwardwarp the flow from $I_{t-1}$ to $I_{t}$ to match the spatial coordinates of corresponding pixels.
When fusing the two flows, Ren~\etal~use an extra network that inputs the flows with their brightness error and outputs the refined final flow.
The underlying idea of inputting the brightness error together is to guide regions to refine to where optical flow may be inaccurate.
In their experiments, Ren~\etal~\cite{chJH:Ren:2019:FAM} demonstrated that utilizing two adjacent optical flows and fusing them improves the flow accuracy especially in occluded areas and out-of-bound areas.

Maurer~\etal~\cite{chJH:Maurer:2018:PFL} also proposed a multi-frame optical flow method that exploits the temporal coherence but in a different direction by learning to predict forward flow from the backward flow in an online manner.
Similarly given three temporally consecutive frames, $I_{t-1}$, $I_{t}$, and $I_{t+1}$, the proposed method first estimates the forward flow (\ie,~from $I_{t}$ to $I_{t+1}$) and the backward flow (\ie,~from $I_{t}$ to $I_{t-1}$) using an off-the-shelf energy-based approach \cite{chJH:Hu:2016:ECF}. 
Next, the method finds inliers for each flow by estimating the opposite directions of each flow (\ie,~from $I_{t+1}$ to $I_{t}$ and from $I_{t-1}$ to $I_{t}$) and performing a consistency check. 
Given the inlier flow for both directions as ground truth data, the method then trains shallow 3-layer CNNs that predict the forward flow (\ie,~from $I_{t}$ to $I_{t+1}$) from the input backward flow (\ie,~from $I_{t}$ to $I_{t-1}$). 
The idea to predict the forward flow from the backward flow is to exploit the valuable motion information from the previous time step including in occluded regions, which the current step is not able to properly handle but that are visible in the previous time step.
This training is done in an online manner so that the network can be trained adaptively to input samples while exploiting temporal coherence.
Finally, the method fuses the predicted forward flow and the estimated forward flow to obtain a refined forward flow.
On major benchmark datasets, the method demonstrates the advantages of exploiting temporal coherence by improving the accuracy especially in occluded regions by up to 27\% overall over a baseline model that does not use temporal coherence.

Finally, Neoral~\etal~\cite{chJH:Neoral:2018:COO} proposed an extended version of PWC-Net \cite{chJH:Sun:2018:PCO} in the multi-frame setting, jointly estimating optical flow and occlusion.
Given a temporal sequence of frames, Neoral~\etal~proposed to improve the flow and occlusion accuracy by leveraging each other in a recursive manner in the temporal domain. 
First, they propose a sequential estimation of optical flow and occlusion: estimating occlusion first and then estimating optical flow, feeding the estimated occlusion as one of inputs into the flow decoder.
They found that providing the estimated occlusion as an additional input improves the flow accuracy by more than 25\%. 
Second, they input the estimated flow from the previous time step into the occlusion and flow decoders as well, which yields additional accuracy improvements for both tasks, especially improving the flow accuracy by more than 12\% on public benchmark datasets.
Similar to other multi-frame based methods above, the flow accuracy improvement is especially prominent in occluded areas and also near motion boundaries.

%% file: chJH_conclusion.tex
\section{Conclusion}
\label{chJH_sec:conclusion}

The recent advances in deep learning have significantly influenced the transition from classical energy-based formulations to CNN-based approaches for optical flow estimation.
We reviewed this transition here.
Two main families of CNN approaches to optical flow have been pursued: \emph{(i)} using CNNs as a feature extractor on top of conventional energy-based formulations and \emph{(ii)} end-to-end trainable, regression-based CNN architectures.
While methods proposed in the initial stages of this transition were outperformed by classical energy-based formulations at the time, steady research progress, \eg~discovering better backbone architectures, synthetic training datasets, and learning strategies eventually led CNN-based methods to yield the most accurate results today and to dominate the current literature.
To overcome the (domain) overfitting tendency of supervised learning, unsupervised or self-supervised methods, as well as semi-supervised learning methods have been recently investigated as alternatives.
Finally, multi-frame CNN approaches, exploiting temporal smoothness or coherency, have demonstrated the potential of improving the flow estimation accuracy even further.

Despite the significant progress, a number of limitations of current approaches remain including, \eg,~\emph{(i)} the domain overfitting tendency, \ie~trained models do not generalize well to unseen domains yet, and \emph{(ii)}
the necessity of complex training schemes, which require pre-training on synthetic datasets first before fine-tuning on the target domain and can make training models complicated in practice.
These and other challenges leave significant room for future work on deep learning methods for optical flow.